\documentclass[lettersize,journal]{IEEEtran}
\usepackage{amsmath,amsfonts}
\usepackage{algorithmic}
\usepackage{algorithm}
\usepackage{array}
\usepackage[caption=false,font=normalsize,labelfont=sf,textfont=sf]{subfig}
\usepackage{textcomp}
\usepackage{stfloats}
\usepackage{url}
\usepackage{verbatim}
\usepackage{makecell}
\usepackage{graphicx}
\usepackage{cite}

\usepackage{xcolor,multirow,graphicx,booktabs}
\usepackage[table]{xcolor}
\begin{document}

\IEEEpubid{
\begin{minipage}{\textwidth}
\scriptsize
Copyright \textcopyright{} 2026 IEEE. Personal use of this material is permitted. However, permission to use this material for any other purposes must be obtained\\
from the IEEE by sending an email to pubs-permissions@ieee.org.
\end{minipage}
}

\title{FSCE: A Target-Aware Frequency-Spatial Collaborative Enhancement Framework for Noise-Resilient SAR ATR}

\author{Yansong Lin,
        Zihan Cheng,
        Ziyue Yang,
        Xinming Wang,
        Jielei Wang,
        Guoming Lu$^{\ast}$,
        Zongyong Cui
\thanks{$^{\ast}$Corresponding author}
\thanks{Y. Lin, Z. Cheng, J. Wang and G. Lu are with the Laboratory of Intelligent Collaborative Computing, University of Electronic Science and Technology of China, China (yansonglin@std.uestc.edu.cn, zihancheng@std.uestc.edu.cn, jieleiwang\_uestc@163.com, lugm@uestc.edu.cn); X. Wang is with the Institute of Automation, Chinese Academy of Sciences, China (wangxinming2024@ia.ac.cn); Z. Yang is with the School of Computer Science and Engineering (School of Artificial Intelligence), Wuhan Institute of Technology, China (ziyueyang@stu.wit.edu.cn); Z. Cui is with the School of Information and Communication Engineering, University of Electronic Science and Technology of China, China (zycui@uestc.edu.cn).}
}



\maketitle

\begin{abstract}

Synthetic aperture radar automatic target recognition (SAR ATR) is severely challenged by coherent speckle noise, whose interference can be progressively amplified by hierarchical nonlinear transformations and eventually damage high-level semantic representations. To address this issue, we propose a Target-Aware Frequency-Spatial Collaborative Enhancement (FSCE) framework for noise-resilient SAR ATR, which integrates frequency-spatial modeling for early feature stabilization with semantic regularization. Specifically, we design a Frequency-Spatial Early-stage Adaptive Enhancement (FS-EAE) module at the network entrance to suppress noise propagation and preserve target structures through collaborative spatial-frequency modeling. Building upon stabilized shallow representation, we further introduce an Adaptive Policy-driven Semantic Alignment (APSA) mechanism, which uses an online teacher policy to impose top-down semantic constraints on the student and feeds semantic guidance back to the enhanced early features during training.
Experiments on MSTAR, OpenSARShip, and FUSARShip demonstrate the effectiveness of this synergy. Moreover, the competitive performance of our lightweight impletation $\text{FSCE-Net}_\mu$ with only 0.17M parameters suggests that the proposed framework is applicable to both high-capacity and lightweight architectures.

\end{abstract}


\section{Introduction}
Synthetic aperture radar (SAR) plays a critical role in domains~\cite{cigna2024exploiting,sunantha2025machine,li2025video} such as geological exploration~\cite{wei2024displacements,chen2025mdca}, maritime surveillance~\cite{elyouncha2024synergistic}, and disaster assessment~\cite{huang2024waterdetectionnet}, owing to its all-weather and day-and-night imaging capabilities~\cite{jiao2025glacier,bai2025investigation}. Despite these advantages, SAR automatic target recognition (ATR) faces formidable challenges rooted in its unique imaging physics~\cite{maharjan2026amplitude, huang2025physics}. On the one hand, coherent signal interference during image formation induces inherent speckle noise, which manifests as stochastic high-frequency fluctuations that obscure fine-grained textures and introduce spurious features. On the other hand, the low contrast between targets and complex backgrounds is often exacerbated by variations in imaging geometry and environmental conditions, significantly complicating the extraction of discriminative structural descriptors.

To mitigate these issues, prior research has explored super-resolution~\cite{wang2025rethinking,dong2025complex}, attributed scattering center (ASC) modeling~\cite{li2021multiscale,feng2022electromagnetic}, and physical properties modeling~\cite{huang2021physics,qin2024scattering}. However, these paradigms exhibit inherent limitations in complex, dynamic environments. SR-based methods often prioritize global pixel-level reconstruction over target-specific feature saliency, while ASC and physical models rely heavily on handcrafted priors and struggle to capture the contextual interplay between targets and their surroundings. Most critically, standard deep learning architectures often ignore the progressive amplification of noise through hierarchical non-linearities, leading to feature fragmentation in deeper semantic layers. Consequently, there is an urgent need for a paradigm that can simultaneously stabilize low-level features and regularize high-level semantic evolution.

By contrast, frequency–spatial coupled modeling exhibits greater flexibility and robustness in feature representation~\cite{miao2024time,wang2025forgotten}. Frequency-domain information emphasizes the response of structural mutation areas and anisotropic scattering centers, whereas spatial-domain representations preserve geometric morphology and contextual relationships. Coupling these two domains enables the complementary exploitation of fine-grained details and global target structure, yielding more discriminative feature representations in complex environments.

Building upon this synergy, we propose a Target-Aware Frequency-Spatial Collaborative Enhancement (FSCE) framework to integrate between early-stage feature filtering and high-level semantic alignment. Unlike existing methods that apply enhancement at arbitrary stages, FSCE introduces a Frequency-Spatial Early-stage Adaptive Enhancement (FS-EAE) module at the network's entry level. This strategically established barrier decouples target signatures from noise-dominant components in both spatial and frequency domains before noise undergoes non-linear amplification. To further regularize the learning process, we introduce an Adaptive Policy-driven Semantic Alignment (APSA) mechanism. Instead of relying on static distillation, APSA utilizes the teacher network to provide adaptive guidance during training, encouraging the student model to learn more robust and discriminative representations. This complementary design enables the model to better handle noise interference and stabilize feature learning across different layers. From a signal-processing perspective, FSCE decouples structural information from noise-related disturbances to preserve discriminative target cues, providing a potentially transferable paradigm for noise-robust target perception.

\IEEEpubidadjcol

The main contributions of this work are summarized as follows:

\begin{enumerate}
\item We propose a Target-Aware Frequency-Spatial Collaborative Enhancement (FSCE) framework for SAR ATR that decouples structural information from noise-related disturbances through early-stage dual-domain enhancement and semantic guidance, providing a potentially transferable principle for noise-robust target-aware perception.
\item We design a unified architecture that combines a Frequency-Spatial Early-stage Adaptive Enhancement (FS-EAE) module for suppressing speckle noise propagation at shallow layers with an Adaptive Policy-driven Semantic Alignment (APSA) mechanism that leverages teacher guidance to improve the robustness and discriminability of learned representations.
\item We demonstrate the effectiveness and generalizability of the proposed framework by instantiating it on different backbone capacities. The full model achieves strong performance on multiple benchmark datasets, while the lightweight model maintains competitive accuracy with only 0.17M parameters, showing strong potential for efficient edge deployment.
\end{enumerate}

\section{Related Work}

\subsection{SAR ATR}

Early research in SAR ATR primarily focused on handcrafted feature extraction combined with shallow classifiers~\cite{cheng2014sar, lin2013optimizing}. These approaches, however, exhibited limited generalization and significant performance degradation under target pose variations. With the emergence of deep learning, convolutional neural networks (CNNs) have redefined the field by enabling end-to-end feature learning. For instance, A-ConvNets \cite{chen2016target} utilized a fully convolutional architecture to mitigate overfitting on the MSTAR dataset, while recent advancements have integrated Transformers \cite{zhao2022few} and domain adaptation \cite{kim2024synthetic} to address data scarcity and simulation-to-measured distribution gaps.

Beyond purely architectural designs, several studies have integrated electromagnetic scattering priors into CNNs. Approaches such as ASC-RISE \cite{gao2024asc} and two-stream graph networks \cite{qin2024scattering} incorporate Attributed Scattering Center (ASC) theory to provide physically interpretable decision features. Although these methods enhance robustness under typical operating conditions, their reliance on explicit scattering models makes them susceptible to background clutter and unmodeled scattering components in complex, real-world environments.

Since the pioneering integration of wavelet transforms into remote sensing by Ranchin and Wald~\cite{ranchin1993wavelet}, multi-resolution analysis has been established as a robust paradigm for SAR speckle suppression and information reconstruction. Building upon this, recent frameworks such as FFDC-Net~\cite{song2023fourier} and WaveCNN-CR~\cite{zi2023wavelet} have successfully leveraged frequency-domain descriptors for crop classification and cloud removal. Nevertheless, these methods typically treat frequency-domain components in isolation or employ static modeling that lacks dynamic interaction with spatial features~\cite{he2019lifting,dang2024dctransformer,li20253d}. Such a decoupled approach offers limited capacity to handle the non-stationary nature of speckle noise, which often results in a trade-off between noise reduction and structural preservation. Consequently, there is a critical need for a unified framework that couples multi-scale spatial fusion with global frequency modulation to maintain structural integrity while achieving adaptive noise suppression.

\subsection{Knowledge Distillation}

With the rapid advancement of deep learning in remote sensing target recognition~\cite{wang2025hitchhiker,liu2026atrnet,wang2026beyond}, model complexity and computational demands have increased accordingly~\cite{wang2025mr,guo2026baddenoise}. To reduce model complexity while maintaining high recognition accuracy, knowledge distillation (KD) technology was widely adopted for model compression and performance enhancement. Traditional knowledge distillation methods typically follow an offline paradigm, in which a high-capacity teacher model is first trained and subsequently used to guide the student model through output supervision. However, offline distillation faces inherent limitations, including fixed teacher representations and inflexible training processes. 

To overcome this, online KD paradigms such as Mutual Contrastive Learning (MCL) \cite{yang2023online} and collaborative learning \cite{guo2020online} enable models to evolve synchronously, improving generalization and training efficiency. In the field of remote sensing target recognition, knowledge distillation also shows great potential. Song \emph{et al.}~\cite{song2023efficient} proposed the KDE-Net model, achieving 88\% parameter reduction while maintaining high accuracy through logit-based KD technology, significantly improving the efficiency of remote sensing image classification. In addition, L\^e and Pham \emph{et al.}~\cite{le2023knowledge} applied KD to the target detection task of remote sensing images, evaluated the performance of various knowledge distillation methods on xView and VEDAI datasets, and verified the effectiveness of KD in remote sensing target detection.

Despite its potential, the application of KD for noise-resilient guidance in SAR ATR remains under-explored. Existing strategies predominantly  focus on accuracy, often neglecting the potential of leveraging a teacher's semantic stability as a structured prior to regularize the student's early-stage feature extraction.

\begin{figure*}[htbp]
  \centering
  \includegraphics[width=0.93\textwidth]{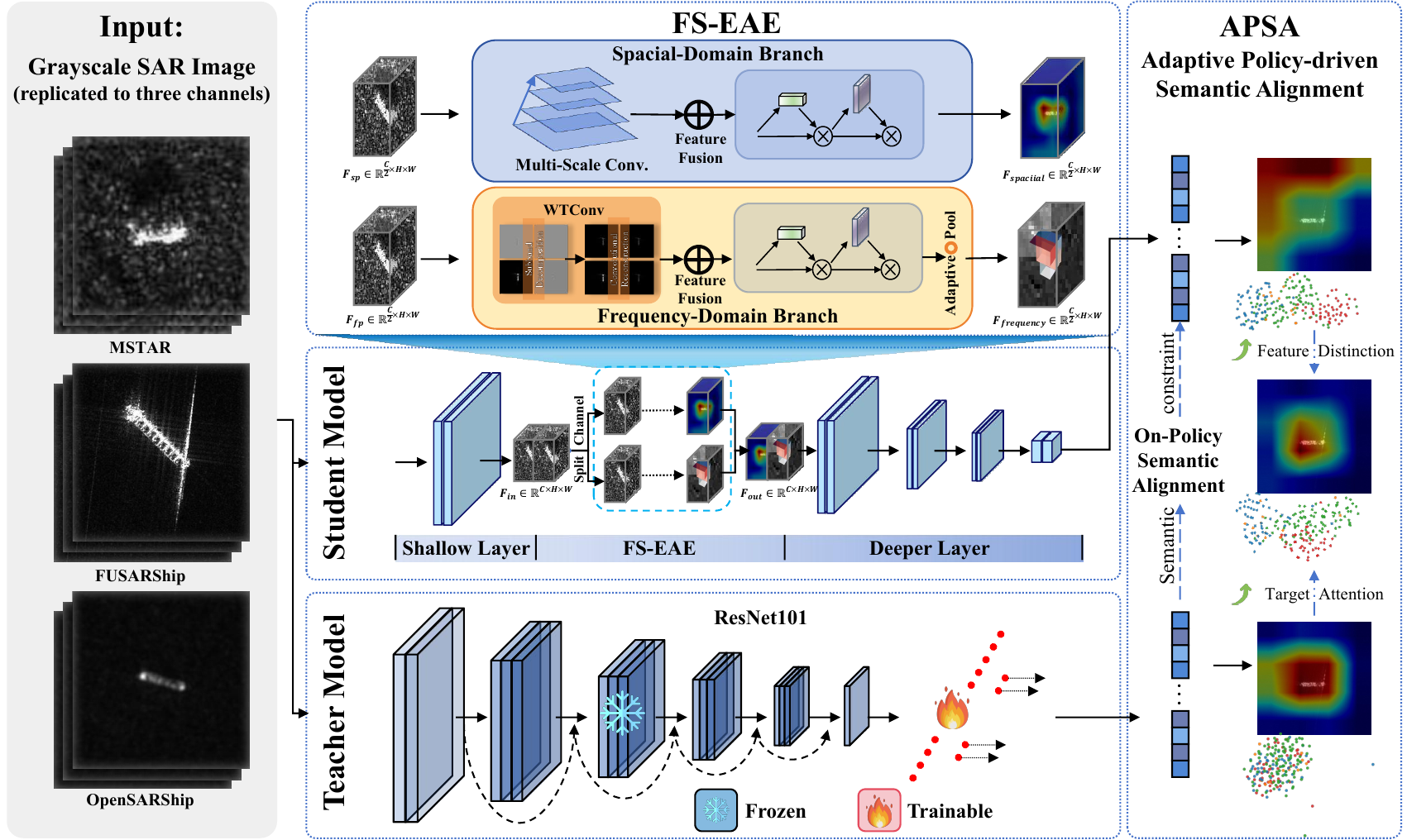}
  \caption{The overall architecture of our proposed method. The framework integrates the FS-EAE module at the early-stage of the student model for dual-domain pre-filtering, which is strategically coupled with the APSA mechanism. The implementation path of each part is given.}
  \label{fig:Model_all}
\end{figure*}

\section{The Proposed Method}

\subsection{Overview of the Proposed Framework}

Given an input SAR grayscale image expanded to three channels to accommodate the input requirements of mainstream networks, the goal of this work is to learn a robust mapping that can effectively suppress speckle noise while preserving discriminative target structures. To this end, we propose a Target-Aware Frequency-Spatial Collaborative Enhancement (FSCE) framework with knowledge-guided noise suppression.

As illustrated in Fig.~\ref{fig:Model_all}, the FSCE framework comprises two complementary components: the Frequency-Spatial Early-stage Adaptive Enhancement (FS-EAE) module and the Adaptive Policy-driven Semantic Alignment (APSA) mechanism. The former is a feature-level component designed to decouple noise-dominant components from signal-relevant structures at the entry level of the network, thereby preventing the non-linear propagation of speckle noise into deeper semantic layers. In parallel, the APSA mechanism provides a learning-level regularization strategy by leveraging a high-capacity teacher network as a semantic guidance source. It introduces a top-down semantic constraint that guides feature learning toward more discriminative representations under noisy conditions.

The proposed framework integrates feature-level structural stabilization and policy-level semantic alignment in a cohesive manner. While FS-EAE stabilizes early-stage feature representations, APSA regularizes representation learning through high-level semantic supervision. This synergy ensures robust SAR target recognition even under severe speckle interference. In the following sections, we detail the architectural design of the FS-EAE module and the APSA mechanism.

\subsection{Frequency-Spatial Early-stage Adaptive Enhancement}

To address the entanglement between speckle noise and structural features, the input feature map $X$ is split along the channel dimension into spatial and frequency subspaces:
\begin{equation}
X \rightarrow \{F_{\text{sp}}, F_{\text{fp}}\}, \quad F_{\text{sp}}, F_{\text{fp}} \in \mathbb{R}^{B \times \frac{C}{2} \times H \times W}.
\end{equation}

We adopt an equal channel split, i.e., a $1{:}1$ ratio, for the spatial and frequency branches. This symmetric allocation provides the two domains with comparable representation capacity without introducing a prior preference for either branch. Moreover, the two $C/2$-channel outputs can be directly concatenated to recover the original channel dimension $C$, avoiding additional projection layers and parameter overhead.

\subsubsection{\textbf{Spatial Multi-Scale Representation}}

The spatial branch $\mathcal{B}_{sp}$ is dedicated to preserving geometric structures under complex scattering conditions. Considering the anisotropic nature of SAR targets, we model multi-scale spatial dependencies using parallel convolutions with varying receptive fields $\mathcal{S}=\{3,5,7,9\}$:
\begin{equation}
F_i =
\begin{cases}
\text{Conv}{3 \times 3}(F_{\text{sp}}), & i = 3, \\
\text{AdaptiveAvgPool}(\text{Conv}_{i \times i}(F{\text{sp}})), & i \in {5,7,9}.
\end{cases}
\end{equation}

As detailed in Eq. (2)-(3), the multi-scale features are aggregated via depthwise separable fusion to ensure computational efficiency while maintaining representational capacity. To further suppress background clutter and enhance target saliency, the Convolutional Block Attention Module (CBAM) is incorporated as a spatial refiner:
\begin{equation}
F_{\text{spatial}} = M_s\big(M_c(F_{\text{agg}}) \odot F_{\text{agg}}\big) \odot \big(M_c(F_{\text{agg}}) \odot F_{\text{agg}}\big).
\end{equation}

\subsubsection{\textbf{Frequency-aware Global Modulation}}

The frequency branch $\mathcal{B}_{fp}$ is designed to capture global scattering characteristics. Unlike spatial dense representations, SAR backscattering is inherently statistical; thus, we leverage frequency-domain analysis to stabilize features.

A Discrete Wavelet Transform (DWT) is first applied to $F_{\text{fp}}$ to isolate hierarchical scattering components:

\begin{equation}
\tilde{F} = \left[ F_{\text{fp}} * \phi_{LL}, F_{\text{fp}} * \phi_{LH}, F_{\text{fp}} * \phi_{HL}, F_{\text{fp}} * \phi_{HH} \right] \downarrow 2,
\end{equation}

where $\{\phi\}$ denotes the Haar wavelet filters used throughout the experiments. The resulting coefficients are encoded into a global scattering prior $z$ through Global Average Pooling (GAP):

\begin{equation}
z = \text{GAP}(\text{CBAM}(\text{ReLU}(\text{BN}(\text{Conv}(\tilde{F}))))).
\end{equation}

Here, $\operatorname{Conv}(\cdot)$ denotes the $5\times5$ depthwise wavelet-domain convolution and inverse reconstruction in WTConv, which maps the concatenated four-subband feature with $2C$ channels back to $\mathbb{R}^{B\times C/2\times H\times W}$. The pooled descriptor $z$ is then broadcast and concatenated with the spatial features.

\subsubsection{\textbf{Unified Representation and Theoretical Filtering}}

The final output of FS-EAE integrates both branches: $F_{\text{FS-EAE}} = \text{Concat}(F_{\text{frequency}}, F_{\text{spatial}})$. 

The comprehensive implementation logic is formalized in Alg.~\ref{alg:fs_eae}.

\begin{algorithm}[H]
\caption{Frequency-Spatial Early-stage Adaptive Enhancement (FS-EAE)}
\label{alg:fs_eae}
\begin{algorithmic}[1]
\REQUIRE Input feature map $X \in \mathbb{R}^{C \times H \times W}$
\ENSURE Enhanced and stabilized feature map $F_{out} \in \mathbb{R}^{C \times H \times W}$

\STATE \textbf{Step 1: Subspace Decomposition}
\STATE $F_{sp}, F_{fp} \leftarrow \text{Split}(X, \text{dim}=1)$ \COMMENT{Decouple spatial and frequency components.}

\STATE \textbf{Step 2(1): Multi-Scale Spatial Modeling}
\FOR{each $k \in \{3, 5, 7, 9\}$}
\STATE $F_{k} \leftarrow \text{ExtractFeature}(F_{sp}, \text{kernel}=k)$
\ENDFOR
\STATE $F_{spatial} \leftarrow \text{CBAM}(\text{Fusion}(F_3, F_5, F_7, F_9))$

\STATE \textbf{Step 2(2): Frequency-aware Modulation}
\STATE $\tilde{F}_{fp} \leftarrow \text{WTConv2d}(F_{fp})$ \COMMENT{Capture statistical scattering responses.}
\STATE $z \leftarrow \text{GlobalAvgPool}(\text{CBAM}(\text{ReLU}(\text{BN}(\tilde{F}_{fp}))))$
\STATE $F_{frequency} \leftarrow \text{Broadcast}(z, \text{target\_size}=(H, W))$

\STATE \textbf{Step 3: Synergistic Integration}
\STATE $F_{out} \leftarrow \text{Concat}(F_{spatial}, F_{frequency}, \text{dim}=1)$
\RETURN $F_{out}$
\end{algorithmic}
\end{algorithm}

To theoretically justify the early-stage placement of FS-EAE, we consider a generic non-linear transformation $F_{l+1} = \sigma(\mathcal{W}_l * F_l + \eta)$. The perturbation $\eta$ propagates as:



\begin{equation}
\sigma(W*F + \eta) \approx \sigma(W*F) + \sigma'(W*F)\cdot \eta,
\end{equation}
where $\sigma'(\cdot)$ acts as a noise gain factor. By applying FS-EAE as a filtering operator $\mathcal{H}_{\text{FS-EAE}}$, we establish:

\begin{equation}
F_{\text{filtered}} = \mathcal{H}_{\text{FS-EAE}}(W_{\text{in}} * X + \eta) \approx W_{\text{in}} * X + \epsilon,
\end{equation}
ensuring $\|\epsilon\| \ll \|\eta\|$. By suppressing stochastic fluctuations at the input level, FS-EAE ensures that the latent representations are sufficiently stable. This structural stabilization provides a clean manifold for the subsequent APSA.

\begin{figure}[h]
    \centering
    \includegraphics[width=0.45\textwidth]{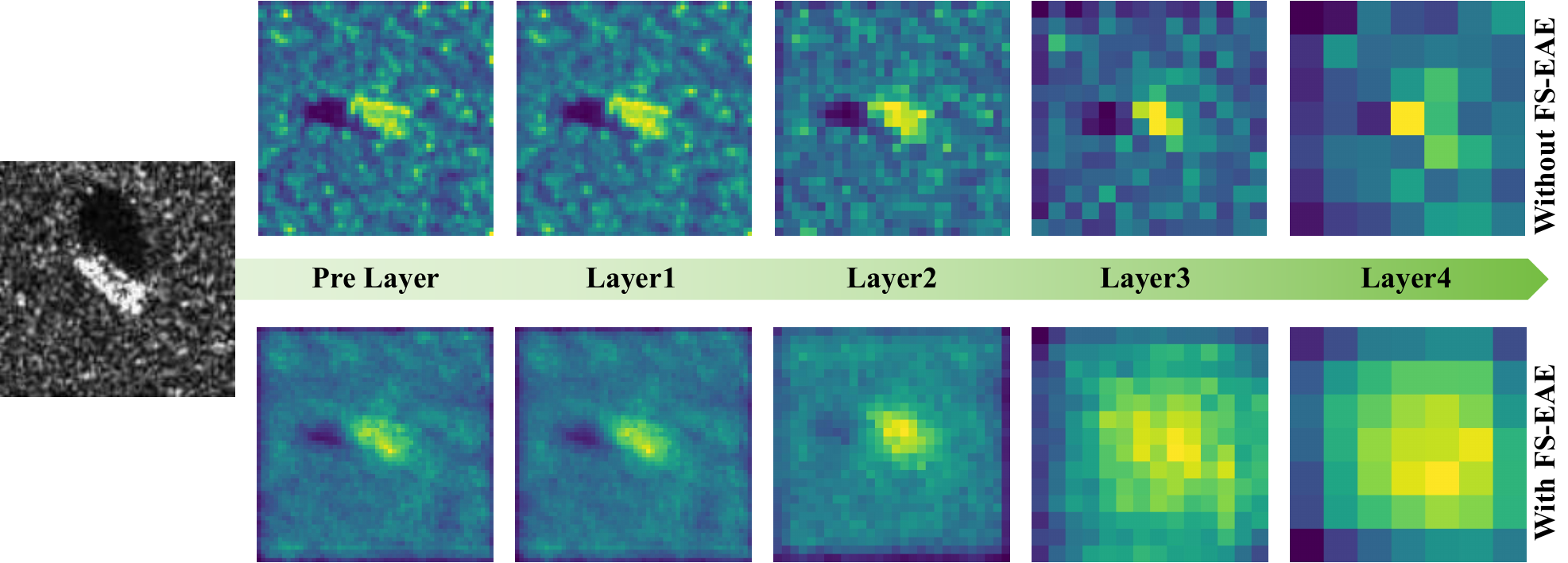}
    \caption{Visualization of feature maps at different depths for ResNet18 and their enhanced versions after FS-EAE.}
    \label{fig:featuremap}
\end{figure}

\begin{figure}[h]
    \centering
    \includegraphics[width=0.45\textwidth]{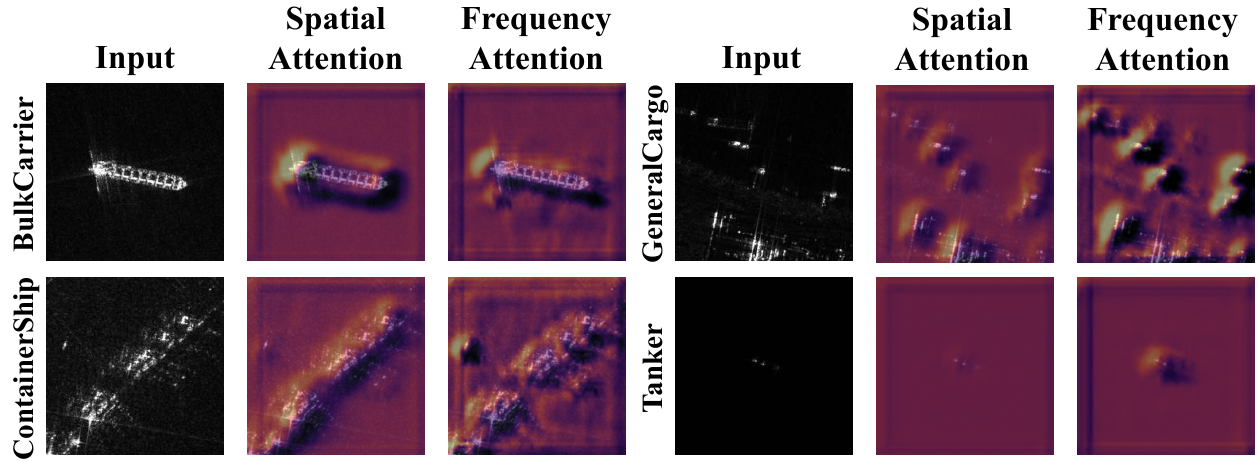}
    \caption{Visualization of spatial- and frequency-branch attention across four ship categories. Columns show the input, spatial attention, and frequency attention, respectively. Warmer colors indicate stronger responses.}
    \label{fig:branch_attention}
\end{figure}

As shown in Fig.~\ref{fig:featuremap}, FS-EAE preserves structural integrity in early-stage features, providing a stable representation that benefits the subsequent hierarchical learning and semantic alignment.

Furthermore, spatial convolution exploits neighborhood correlations to capture coherent contours, but weak scattering points occupying only a few pixels may be diluted by surrounding background responses. Wavelet decomposition preserves such sparse responses through localized frequency subbands, although it provides less geometric context and may react to high-frequency clutter. FS-EAE therefore combines spatial continuity with localized spectral evidence. As shown in Fig.~\ref{fig:branch_attention}, the spatial branch emphasizes the extended BulkCarrier structure but produces a more confined response for the small Tanker, while the frequency branch retains a salient response around its sparse scattering point.

\subsection{Adaptive Policy-driven Semantic Alignment}

SAR imagery is inherently affected by coherent speckle noise, which introduces high-frequency stochastic fluctuations that are largely irrelevant to target semantics. Due to hierarchical nonlinear transformations in deep networks, such noise can be progressively amplified, leading to feature fragmentation and degraded discriminability in deeper semantic layers.

To mitigate this propagation, we propose the Adaptive Policy-driven Semantic Alignment (APSA) mechanism. The core idea is to leverage the semantic output of a high-capacity teacher network as a dynamic guidance signal to regularize the early-stage features produced by the FS-EAE module. This introduces a top-down semantic constraint that encourages the student network to learn target-consistent representations in the presence of speckle noise.

\subsubsection{\textbf{Synergistic Online Policy Alignment}}


In APSA, the teacher uses an ImageNet-pretrained ResNet101 backbone with a randomly initialized task-specific classification head. During training, the backbone parameters remain fixed, while only the classifier head is updated online using cross-entropy loss to adapt to the SAR data distribution.

Specifically, this process can be interpreted as encouraging the student to match the teacher’s predictive distribution, thereby improving the alignment between learned features and ground-truth semantics. We minimize the Kullback-Leibler (KL) divergence between the teacher and student outputs:

\begin{equation}
\mathcal{L}_{\text{KD}} = T^2 \cdot \mathrm{KL}(p_t \parallel p_s),
\end{equation}
where $p_t$ and $p_s$ denote the teacher and student prediction distributions, respectively, and $T$ controls the softness of the probability distribution. In addition to the distillation loss, the student is optimized with the standard cross-entropy loss. The overall objective is:

\begin{equation}
\mathcal{L}_{s} = \mathcal{L}_{\text{CE}} + \alpha \mathcal{L}_{\text{KD}},
\end{equation}
where $\alpha$ balances the two terms.

Through this online updating strategy, the teacher gradually adapts to the target domain while maintaining stable semantic guidance for the student network. This is particularly beneficial under noisy SAR conditions.

\begin{figure}[b]
    \centering
    \includegraphics[width=0.45\textwidth]{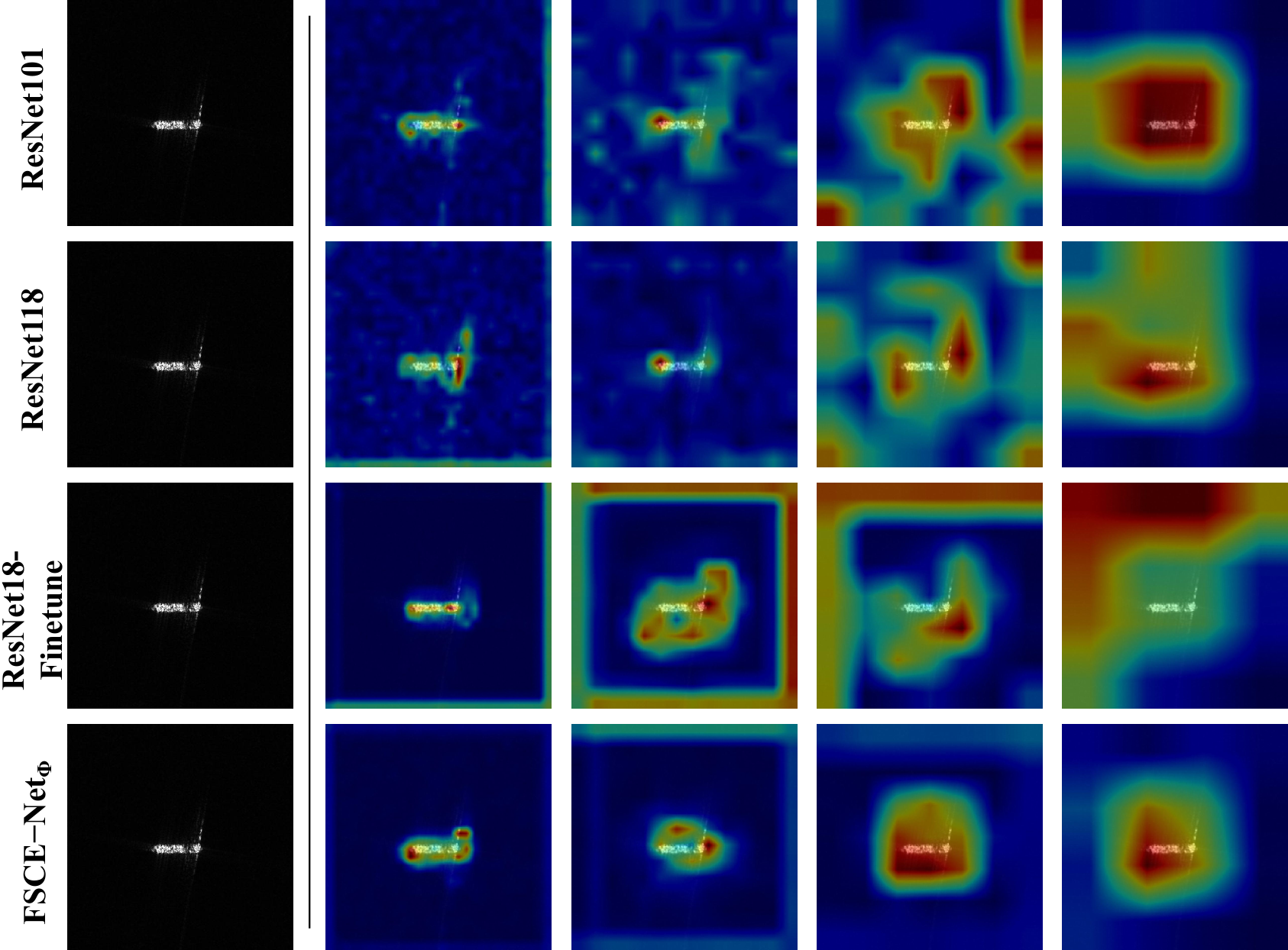} 
    \caption{Grad-CAM is used to compare the heat maps of the interest areas of different layers of each model. From top to bottom, the models are ResNet101 pre-trained model, ResNet18 pre-trained model, ResNet18 model without pretraining but with the same training configuration as $\text{FSCE-Net}_{\Phi}$, and $\text{FSCE-Net}_{\mu}$.}
    \label{fig:high-frequency}
\end{figure}

\subsubsection{\textbf{Semantic Regularization Effect on FS-EAE Features}}

To explicitly align between abstract semantic supervision and early-stage feature enhancement, we formulate the APSA regularization as a functional dependency acting on the FS-EAE representations:

\begin{equation}
\frac{\partial \mathcal{L}_{\text{KD}}(z_s, z_t)}{\partial f_s^{(l)}} =
\frac{\partial \mathcal{L}_{\text{KD}}}{\partial p_s}
\frac{\partial p_s}{\partial z_s}
\frac{\partial z_s}{\partial f_s^{(l)}}.
\end{equation}
where $z_s$ is produced from the intermediate student representation $f_s^{(l)}$. This indicates that the semantic guidance provided by the teacher is not limited to the classifier output, but also regularizes the feature extraction process of the student network. As a result, the FS-EAE module is encouraged to produce representations that are more consistent with class-discriminative semantic structures and less sensitive to noise-dominant perturbations. Under noisy SAR conditions, such semantic regularization helps suppress unstable responses caused by speckle interference while preserving structurally meaningful target cues.

As illustrated in Fig.~\ref{fig:high-frequency}, the Grad-CAM visualization clarifies the impact of APSA on noise robustness. In the deeper teacher model (ResNet101), the attention maps consistently focus on the primary target, filtering out background interference. By integrating the FS-EAE module and leveraging APSA, $\text{FSCE-Net}_{\Phi}$ achieves a concentration of attention similar to the teacher model. This indicates that the proposed APSA mechanism effectively enhances semantic consistency across network layers and improves robustness under noisy SAR conditions.


\subsection{Model Instantiation and Deployment}

To accommodate different computational constraints in SAR ATR, we instantiate the proposed FSCE framework on backbones with varying capacities:

\begin{itemize}
    \item \textbf{$\text{FSCE-Net}_{\Phi}$:} Built upon the \textbf{ResNet18} backbone, this variant focuses on achieving strong recognition performance in complex scenarios.
    \item \textbf{$\text{FSCE-Net}_{\mu}$:} Utilizing the \textbf{ShuffleNetV2 0.25x} architecture, this variant offers an ultra-lightweight design (0.17M parameters) for real-time deployment on resource-constrained edge devices.
\end{itemize}

The FS-EAE module is inserted at the early stage—after Layer 1 in ResNet18 and after Conv1 in ShuffleNetV2 0.25x. As illustrated in Fig.~\ref{fig:featuremap}, shallow features retain higher spatial resolution and preserve fine-grained structural details. Performing frequency-spatial enhancement at this stage helps suppress speckle noise before it propagates through deeper layers, thereby improving the stability of subsequent feature extraction.

\section{Experiments}

In this section, numerous experiments are conducted and experimental settings are provided to clearly demonstrate the effectiveness of our method. First, three SAR ATR datasets used for experiments are introduced, as well as their similarities and differences. Then, hyperparameter and ablation experiments are conducted to evaluate the effectiveness of the proposed Frequency-Spatial Early-stage Adaptive Enhancement (FS-EAE) module and Adaptive Policy-driven Semantic Alignment (APSA) mechanism. Finally, multiple comparative experiments with different types of methods were conducted to verify the superiority of our method.
\subsection{Dataset Details and Data Preprocessing}\label{AA}

To evaluate the effectiveness of the FSCE framework, three publicly available SAR datasets were employed: MSTAR, OpenSARShip, and FUSARShip.

\subsubsection{Dataset Descriptions}

\begin{figure}[H]
    \centering
    \includegraphics[width=0.45\textwidth]{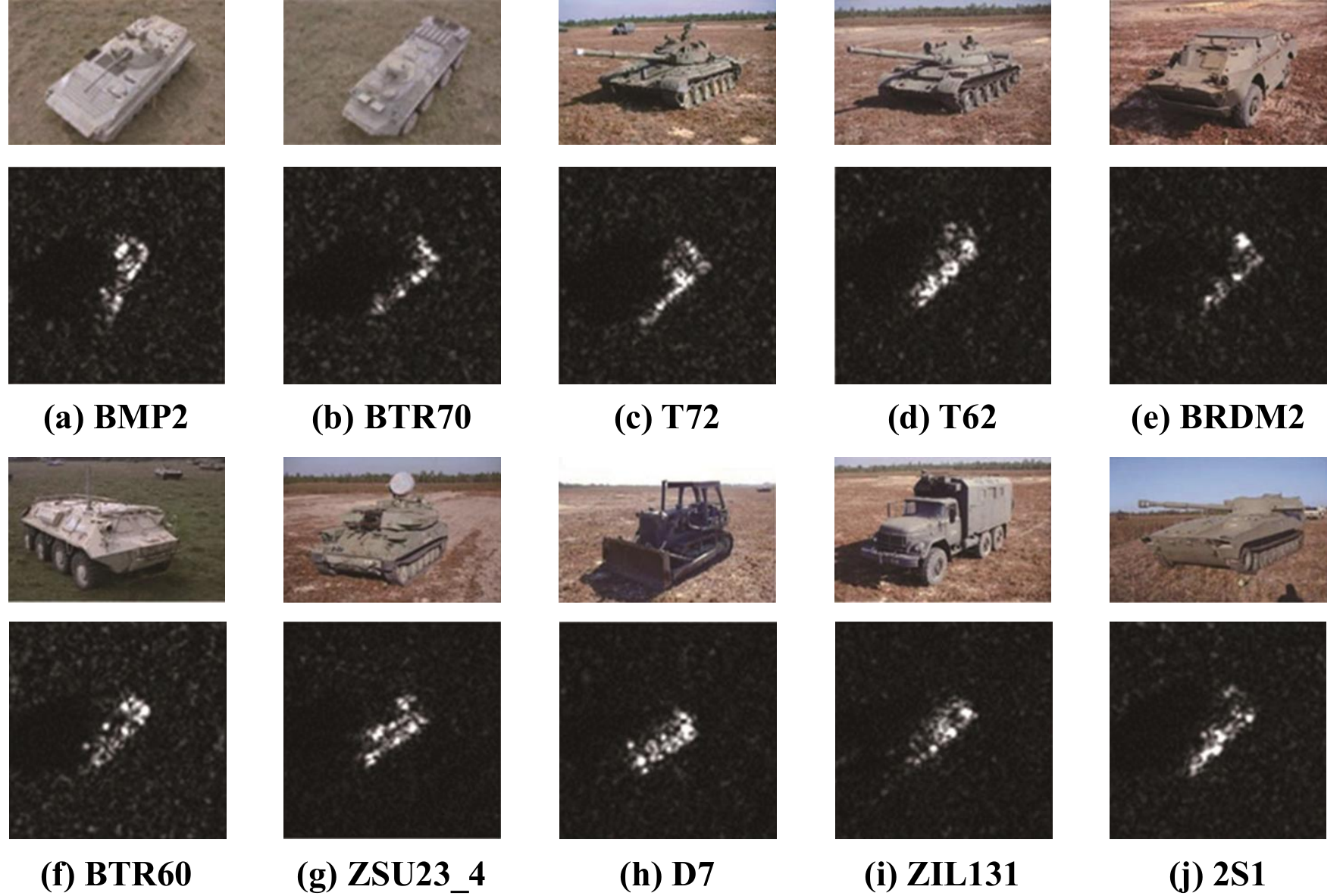}
    \caption{MSTAR 10-classification dataset.}
    \label{fig:MSTAR}
\end{figure}

\textbf{MSTAR Dataset:} The MSTAR dataset~\cite{keydel1996mstar} is a set of SAR images provided by the Defense Advanced Research Project Agency (DARPA) and the Air Force Research Laboratory (AFRL). This dataset collected images of ten different categories of ground military vehicles (armored personnel carriers: BMP2, BRDM2, BTR60, and BTR70; tanks: T62, T72; air defense units: ZSU\_23\_4; bulldozers: D7; rocket launchers: 2S1; and trucks: ZIL131) based on the synthetic aperture radar sensor platform of Sandia National Laboratories, as shown in Fig.~\ref{fig:MSTAR}.

For each category, SAR images were collected with all-around azimuth coverage. The spatial resolution distributions of the training and test sets were summarized in Tab.~\ref{tab:MSTAR}.

\begin{figure}[H]
    \centering
    \includegraphics[width=0.45\textwidth]{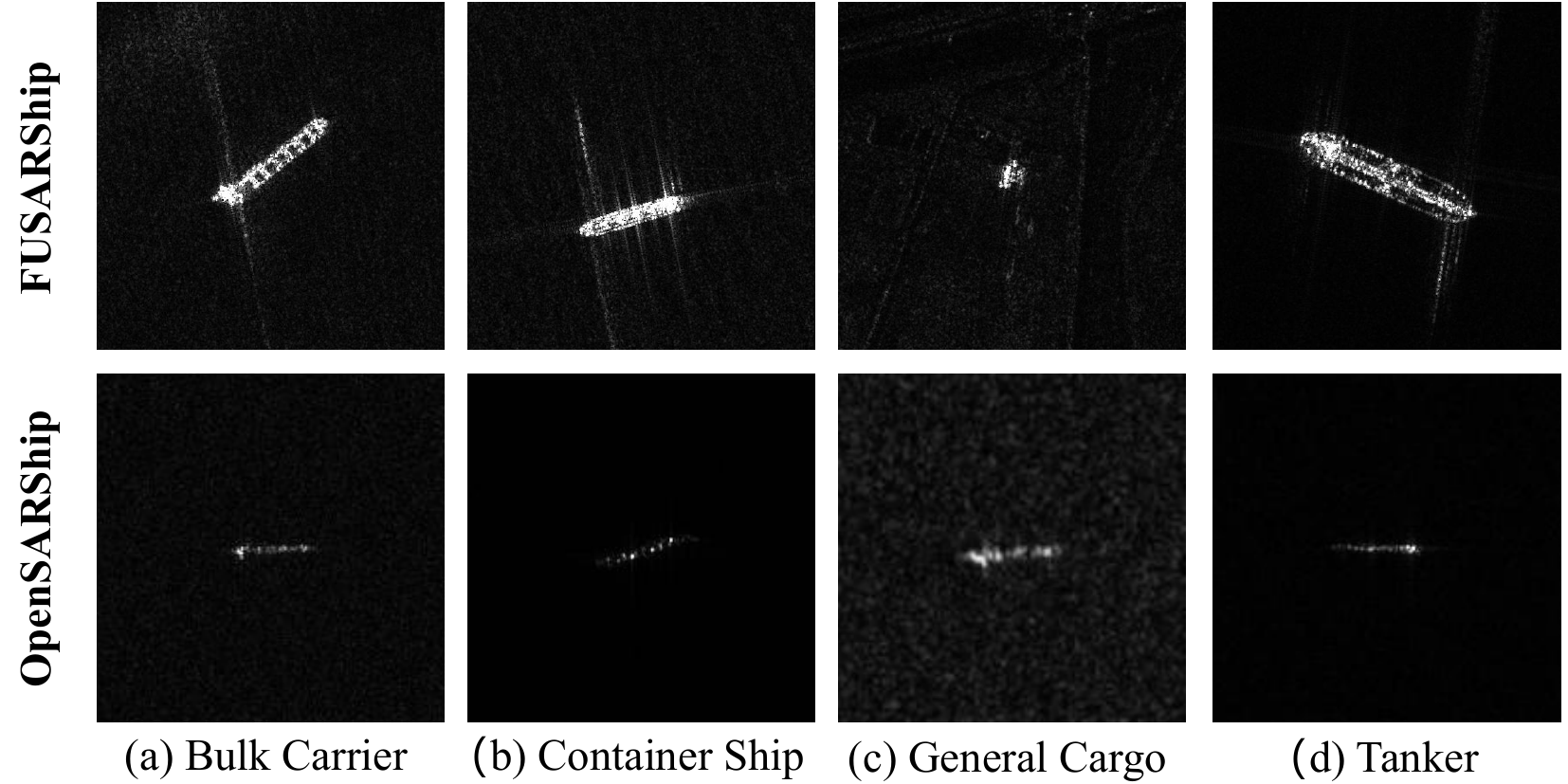}
    \caption{OpenSARShip and FUSARShip 4-classification datasets}
    \label{fig:FUOPEN}
\end{figure}

\begin{table}[H]
\centering
\caption{Statistics of Samples in the MSTAR dataset.}
\label{tab:MSTAR}
\resizebox{0.48\textwidth}{!}{
\normalsize
\begin{tabular}{c|cccc}
\toprule
\multirow{2}{*}{\textbf{Serial No.}} & \multicolumn{2}{c}{\textbf{Training}} & \multicolumn{2}{c}{\textbf{Test}} \\
\cmidrule(lr){2-3} \cmidrule(lr){4-5}  
 & \textbf{Elevation} & \textbf{Number} & \textbf{Elevation} & \textbf{Number} \\ 
\midrule
ZSU\_23\_4       & 17$^{\circ}$ & 299 & 15$^{\circ}$ & 274 \\ 
BRDM\_2        & 17$^{\circ}$ & 298 & 15$^{\circ}$ & 274 \\ 
BTR60          & 17$^{\circ}$ & 256 & 15$^{\circ}$ & 195 \\ 
BTR70     & 17$^{\circ}$ & 233 & 15$^{\circ}$ & 196 \\ 
BMP2      & 17$^{\circ}$ & 233 & 15$^{\circ}$ & 196 \\ 
D7       & 17$^{\circ}$ & 299 & 15$^{\circ}$ & 274 \\ 
ZIL131   & 17$^{\circ}$ & 299 & 15$^{\circ}$ & 274 \\ 
2S1      & 17$^{\circ}$ & 299 & 15$^{\circ}$ & 274 \\  
T62                & 17$^{\circ}$ & 299 & 15$^{\circ}$ & 273 \\ 
T72               & 17$^{\circ}$ & 232 & 15$^{\circ}$ & 196 \\ \hline
Total                      & -            & 2747 &   -          & 2426 \\ 
\bottomrule
\end{tabular}%
}
\end{table}

\textbf{FUSARShip and OpenSARShip Datasets:} The FUSARShip high-resolution SAR ship dataset~\cite{hou2020fusar} was acquired by the Gaofen-3 satellite under dual-polarization modes (DH and DV). It contained 15 major ship categories, 98 subcategories, as well as a variety of non-ship marine targets. The spatial resolution for all images ranged from 1.7-1.754m$\times$1.124m (range$\times$azimuth). Bulk Carrier, Container Ship, General Cargo, and Tanker were selected. The image size was resized to 256$\times$256 in the data processing section, and then the center was cropped to 224$\times$224 pixels for better feature learning and target recognition. Representative samples from the datasets are shown in Fig.~\ref{fig:FUOPEN}.

The OpenSARShip dataset~\cite{li2017opensarship} was acquired by the Sentinel-1 satellite. The dataset contains 17 types of targets in ground range detection (GRD) and single-look complex (SLC) modes. The spatial resolution is 20m$\times$22m (range$\times$azimuth) in GRD mode and 2.7-3.5m$\times$22m (range$\times$azimuth) in SLC mode. The dataset exhibits a highly imbalanced category distribution (e.g., 8,470 cargo ships versus only 214 tugboats). To mitigate the impact of severe class imbalance, we selected Bulk Carrier, Container Ship, General Cargo, and Tanker. Following the same preprocessing pipeline as FUSARShip, the images were resized to 256$\times$256 and then the center-cropped to 224$\times$224 pixels. Representative samples from the dataset are shown in Fig.~\ref{fig:FUOPEN}.

\begin{table}[H]
\centering
\caption{Statistics of Samples in the OpenSARShip and FUSARShip Datasets}
\label{tab:FUOPEN}
\normalsize
\resizebox{0.48\textwidth}{!}{%
\begin{tabular}{c|ccccc}
\toprule
\textbf{Dataset} & \textbf{Bulk Carrier} & \textbf{Container Ship} & \textbf{General Cargo} & \textbf{Tanker} & \textbf{Total} \\
\midrule
OpenSARShip & 910 & 326 & 227 & 647 & 2110 \\
FUSARShip  & 262 & 56  & 464 & 240 & 1022 \\ 
\bottomrule
\end{tabular}%
}
\end{table}

The overall situation of the FUSARShip and the OpenSARShip dataset used in this paper is shown in Tab.~\ref{tab:FUOPEN}.

For the FUSARShip and OpenSARShip datasets, the samples are randomly divided into training and test sets, with 80\% of the data used for training and the remaining 20\% reserved for testing.

\subsubsection{Unified Preprocessing and Data Augmentation}

To ensure consistent input for the FSCE framework, all raw SAR images undergo a standardized preprocessing pipeline. All samples are converted to $224 \times 224$ pixels and normalized before being fed into the network.

Dual-View Augmentation Strategy: To enhance model robustness and facilitate online knowledge distillation, we adopt a dual-view data augmentation strategy during the training phase:

\begin{itemize}
\item \textbf{Teacher-View Augmentation:} The teacher view employs random horizontal and vertical flipping with $p=0.5$, random rotation within $\pm270^{\circ}$, and random affine transformation with a rotation range of $\pm10^{\circ}$ and a maximum translation ratio of $0.1$. Random perspective transformation uses a distortion scale of $0.2$ with $p=0.5$, while random erasing is applied with $p=0.1$ and an erasing-area ratio of $[0.02,0.20]$.
\item \textbf{Student-View Augmentation:} The student view employs random resized cropping to $224\times224$ with a scale range of $[0.8,1.0]$. Gaussian blur with a kernel size of $5$ and $\sigma\in[0.1,2.0]$ is applied with $p=0.5$. Random horizontal and vertical flipping are both applied with $p=0.5$, followed by random rotation within $\pm270^{\circ}$. Adaptive speckle noise is generated using a Gamma distribution with an equivalent number of looks $L=5$.
\end{itemize}

This dual-view approach enables the framework to learn multi-angle and noise-invariant features from the same underlying sample, effectively regularizing the distillation process.

\subsection{Experimental Setup}

During training, the number of iterations was set to 300 with a batch size of 64. The learning rate was initialized to 2.5e-4 for the teacher network and 2.5e-3 for the student network. The teacher network employed a Cosine Annealing learning rate scheduler to gradually decay the learning rate and enhance generalization, while the student network adopted the OneCycle learning rate scheduler, which rapidly increases the learning rate in the early training stage and then gradually decreases it. The AdamW optimizer was used. 

All experiments were conducted using the PyTorch deep learning framework on an NVIDIA GeForce RTX 4060 GPU. The experimental environment was configured with PyTorch 1.13, Python 3.9, CUDA 11.7, and Windows 11.

\begin{table*}[t]
\centering
\caption{Performance comparison of mainstream methods and the proposed $\text{FSCE-Net}$ across multiple SAR datasets. The best results are highlighted in \textbf{bold}, and the second-best results are highlighted with \underline{underline}. $\uparrow$: higher is better; $\downarrow$: lower is better; "--": not reported. $^{\dagger}$ Results are quoted from existing studies. Complexity metrics are measured with a 4-class classification head and an input resolution of $224 \times 224$.}
\label{tab:sar_all_merge_compact}
\renewcommand{\arraystretch}{1.2}
\resizebox{1\textwidth}{!}{%
\begin{tabular}{c|cc|c|c|c}
\toprule
\textbf{Method} & \textbf{Params (M)}$^{\ddagger}$ $\downarrow$ & \textbf{FLOPs (M)}$^{\ddagger}$ $\downarrow$ & \textbf{MSTAR Acc (\%)} $\uparrow$ & \textbf{OpenSARShip Acc (\%)} $\uparrow$ & \textbf{FUSARShip Acc (\%)} $\uparrow$ \\ 
\midrule
\multicolumn{6}{c}{\cellcolor[gray]{1}\textbf{General Deep Learning Models (CNNs / Transformers / Frequency--Spatial Fusion)}} \\
\midrule
EfficientNet-b0~\cite{tan2019efficientnet} & 5.25 & 385.88 & 97.48 & 70.28 & 83.01 \\
Inception-v3~\cite{szegedy2016rethinking} & 23.80 & 2837.97 & 96.53 & 74.29 & 84.95 \\
DenseNet-121~\cite{huang2017densely} & 7.90  & 2834.39  & 94.76 & 75.70 & 83.98 \\
ConvNeXt-v2-A~\cite{woo2023convnext} & 3.69 & 547.35 & 99.37 & 50.47 & \underline{86.41} \\
ROPE-ViT-small~\cite{heo2024rotary} & 21.96 & 4241.22 & 94.48 & 75.00 & 83.98 \\
Twins-B~\cite{chu2021twins} & 43.83 & 6457.16 & 81.88 & \underline{76.18} & 85.44 \\
StarNet-s1~\cite{ma2024rewrite}   & 2.68 & 427.14 & 78.16 & 61.08 & 82.26 \\
VOLO-d1~\cite{yuan2022volo}      & 26.56 & 6517.14 & 97.71 & 75.71 & \underline{86.41} \\   
FFC-ResNet50~\cite{chi2020fast} & 28.14 & 4633.24 & \underline{99.57}  & 75.24 & 84.47 \\
\midrule
\multicolumn{6}{c}{\cellcolor[gray]{1}\textbf{Existing SAR ATR Specific Methods}} \\
\midrule
A-ConvNet$^{\dagger}$~\cite{chen2016target} & \underline{0.304} & 455.19 & 99.13 & 71.05 & 83.93 \\
ESENet$^{\dagger}$~\cite{wang2019sar} & 0.54 & \textbf{295.41} & 97.32 & 73.06 & 82.74 \\
FENNet$^{\dagger}$~\cite{zeng2021unknown} & 0.66 & 584.03 & 97.32 & 71.69 & 84.02 \\
ResNet-DTL~\cite{huang2020classification} & 11.18 & 1823.52 & 99.46$^{\dagger}$ & 72.17 & 82.52 \\
ResNet18+IFTS$^{\dagger}$~\cite{choi2022fusion} & 22.35 & 3490.46 & 98.90 & -- & -- \\
CA-MCNN$^{\dagger}$~\cite{li2021multiscale} & 2.78 & 60389.41 & 97.81 & -- & -- \\
GSP-IF$^{\dagger}$~\cite{jingyi2025target} & 24.97 & 14190.50 & 98.56 & -- & -- \\
PAE-SD$^{\dagger}$~\cite{wang2024multiresolution} & 11.17 & 1770.51 & -- & 75.61 & 85.05 \\
\midrule
\multicolumn{6}{c}{\cellcolor[gray]{1}\textbf{Proposed Methods}} \\
\midrule
\textbf{$\text{FSCE-Net}_{\Phi}$ (Ours)} & 11.40 & 2374.05 & \textbf{99.59} & \textbf{76.42} & \textbf{89.32} \\
\textbf{$\text{FSCE-Net}_{\mu}$ (Ours)} & \textbf{0.17} & \underline{343.87} & 99.37 & 74.76 & \underline{86.41} \\
\bottomrule
\end{tabular}%
}
\end{table*}

\subsection{Comparison with Representative Methods}\label{AA}

To thoroughly evaluate the effectiveness of the proposed framework, we conducted comparative experiments against several representative mainstream methods on three SAR datasets.

\subsubsection{Compared Methods}

The competitive methods selected for comparison can be categorized into four groups~\cite{rw2019timm}:
\begin{itemize}
\item \textbf{General CNN Models:} Including \textbf{Inception-v3}~\cite{szegedy2016rethinking}, \textbf{DenseNet}~\cite{huang2017densely}, and \textbf{EfficientNet}~\cite{tan2019efficientnet}, which are classic architectures widely used in optical image recognition.
\item \textbf{Modern Vision Backbones:} Including \textbf{ConvNeXt-v2}~\cite{woo2023convnext}, \textbf{StarNet}~\cite{ma2024rewrite}, \textbf{VOLO}~\cite{yuan2022volo}, and
\textbf{FFC-ResNet50}~\cite{chi2020fast}. In particular, FFC-ResNet50 is included as a representative generic
frequency--spatial fusion architecture that combines local spatial
convolutions with semi-global and global Fourier processing.


\item \textbf{Transformer-based Models:} Including \textbf{ROPE-ViT}~\cite{heo2024rotary} and \textbf{Twins}~\cite{chu2021twins}, which leverage self-attention mechanisms to capture long-range dependencies.
\item \textbf{SAR-Specific ATR Methods:} Including \textbf{A-ConvNet}~\cite{chen2016target} (a pioneering all-convolutional network replacing fully connected layers to prevent overfitting), \textbf{ESENet}~\cite{wang2019sar} (an enhanced squeeze-and-excitation module for discriminative feature extraction), \textbf{FENNet}~\cite{zeng2021unknown} (a dedicated feature enhancement network for identifying unknown targets), \textbf{ResNet-DTL}~\cite{huang2020classification} (a deep transfer learning paradigm for domain adaptation), \textbf{ResNet18+IFTS}~\cite{choi2022fusion} (synergistic fusion of target and shadow features), \textbf{CA-MCNN}~\cite{li2021multiscale} (a multi-scale architecture for capturing hierarchical scattering centers), \textbf{GSP-IF}~\cite{jingyi2025target} (a graph perception network for preserving invariant topological scattering relationships), and \textbf{PAE-SD}~\cite{wang2024multiresolution} (a combination of physical attention and scale distillation to mitigate resolution variations). To ensure a fair comparison, we strictly utilize the single-domain accuracy reported in the original paper of PAE-SD, as the cross-domain dataset fusion strategy employed in their work.
\end{itemize}

The quantitative results are summarized in Tab.~\ref{tab:sar_all_merge_compact}.
 

\subsubsection{Comparative Analysis and Discussions}

We analyze the experimental results from the following four perspectives:

\paragraph{\textbf{Superiority in Recognition Accuracy}}The proposed $\text{FSCE-Net}_{\Phi}$ achieves superior performance across all three datasets. Specifically, on the challenging FUSARShip dataset, it attains an accuracy of \textbf{89.32\%}, outperforming the tied second-best external methods, ConvNeXt-v2-A and VOLO-d1, by \textbf{2.91} percentage points. On OpenSARShip, it reaches \textbf{76.42\%}, which is 0.24\% higher than the next-best result. On MSTAR, FSCE-Net$_{\Phi}$ achieves \textbf{99.59\%}, slightly surpassing FFC-ResNet50 (99.57\%). It also consistently outperforms the similar-backbone ResNet-DTL and improves over FFC-ResNet50 by 1.18 and 4.85 percentage points on OpenSARShip and FUSARShip, respectively. These results demonstrate the effectiveness of FS-EAE and APSA over conventional transfer learning and generic Fourier-based fusion.

\paragraph{\textbf{Comparison with Frequency--Spatial Fusion}} Compared with FFC-ResNet50, FSCE-Net$_{\phi}$ achieves comparable MSTAR accuracy and gains 1.18 and 4.85 percentage points on OpenSARShip and FUSARShip, with 59.5\% fewer parameters. These larger maritime gains indicate that shallow localized wavelet filtering, spatial modeling, and APSA suppress clutter more effectively than generic global Fourier fusion.

\paragraph{\textbf{Efficiency and Edge Deployment Suitability}}The proposed $\text{FSCE-Net}_{\mu}$ (Efficiency-Centric) demonstrates an exceptional trade-off between complexity and performance. With an extremely compact parameter size of only \textbf{0.17M}, it outperforms many heavy models like Inception-v3 (23.8M) and DenseNet-121 (7.9M). Remarkably, on FUSARShip, $\text{FSCE-Net}_{\mu}$ achieves an accuracy of 86.41\%, matching the performance of VOLO-d1 while using less than \textbf{0.7\%} of its parameters. This lightweight advantage makes our model highly suitable for real-time deployment on resource-constrained edge devices.

\begin{figure}[h]
    \centering
    \includegraphics[width=0.45\textwidth]{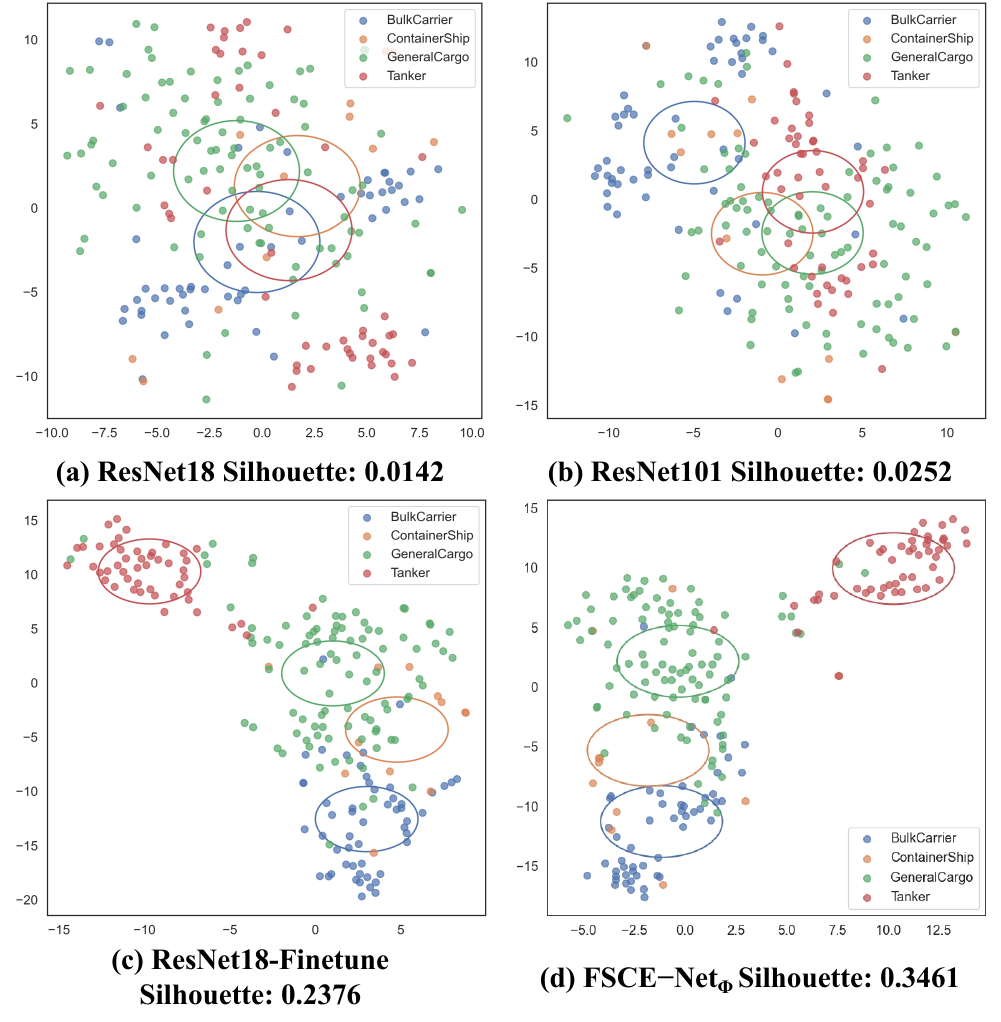}
    \caption{t-SNE visualization and silhouette score on FUSARShip. (a) Pre-trained ResNet18, (b) Pre-trained ResNet101, (c) Fine-tuned ResNet18 without pre-training, (d) $\text{FSCE-Net}_{\Phi}$ (Ours).}
    \label{fig:tsne}
\end{figure}

\paragraph{\textbf{Robustness in Complex Scenarios}}The performance gains are particularly evident in the maritime datasets (OpenSAR and FUSAR), which are characterized by high-intensity sea clutter and varying spatial resolutions. Standard models primarily designed for optical images (e.g., StarNet, Twins-B) lack mechanisms to handle the unique speckle noise of SAR. In contrast, our method adaptively models both frequency and spatial domains, effectively separating high-frequency noise from target structures. While methods like PAE-SD struggle with local blur or occlusion, the APSA mechanism in FSCE-Net ensures that the student network aligns with a stable semantic policy, leading to more discriminative feature clusters. This is further validated by Fig.~\ref{fig:tsne}, where our model achieves the highest silhouette score, indicating superior inter-class separability.

\paragraph{\textbf{Feature Clustering Quality}} 
To further investigate the discriminative power of the learned representations, we employ t-SNE visualization and the silhouette score on the FUSARShip dataset, as shown in Fig.\ref{fig:tsne}. The silhouette score quantifies the clustering effectiveness, where higher values indicate superior inter-class separability and intra-class compactness. 

The visualization results reveal that $\text{FSCE-Net}_{\Phi}$ exhibits the most structured and discriminative feature distribution among the compared models. Specifically, our model achieves the highest silhouette score, marking an improvement of \textbf{0.3319} over the pre-trained ResNet18. This demonstrates that the synergy between FS-EAE and APSA effectively guides the student network to converge toward a more separable semantic manifold, validating the superior representation learning capability of the proposed framework.

\subsection{FSCE Framework Transferability and Generalization}\label{AA}

To evaluate the synergistic effect of the FS-EAE module and the APSA mechanism on model generalization, we conducted transferability experiments across various backbones.

\begin{table}[h]
\centering
\normalsize
\caption{
Accuracy comparison with and without APSA on OpenSARShip and FUSARShip. The best results of each model are highlighted in \textbf{bold}. \(\uparrow\) indicates that a higher value denotes better performance.
}
\label{tab:kd_vertical}
\resizebox{0.48\textwidth}{!}{%
\begin{tabular}{c|ccc}
\toprule
\textbf{Dataset} & \textbf{Model} & \textbf{With APSA (\%)~\(\uparrow\)} &\textbf{Without APSA (\%)~\(\uparrow\)} \\ \hline
\multirow{6}{*}{\rotatebox[origin=c]{90}{OpenSARShip}} 
& A-ConvNet & 65.80 & 62.50 \\
& A-ConvNet + FS-EAE & \textbf{73.11} & 72.17 \\
& ResNet34 & 75.24 & 75.00 \\
& ResNet34 + FS-EAE & \textbf{76.18} & 75.47 \\
& ResNet18 & 75.47 & 72.17 \\
& \textbf{Ours (ResNet18 + FS-EAE)} & \textbf{76.42} & 74.76 \\ 
\midrule

\multirow{6}{*}{\rotatebox[origin=c]{90}{FUSARShip}} 
& A-ConvNet & 81.07 & 77.18 \\
& A-ConvNet + FS-EAE & \textbf{83.50} & 82.52 \\
& ResNet34 & 83.98 & 83.50 \\
& ResNet34 + FS-EAE & \textbf{84.95} & 84.47 \\
& ResNet18 & 83.98 & 83.01 \\
& \textbf{Ours (ResNet18 + FS-EAE)} & \textbf{89.32} & 84.47 \\ 
\bottomrule
\end{tabular}%
}
\end{table}

As shown in Tab.~\ref{tab:kd_vertical}, the integration of FS-EAE consistently yields performance gains. For ResNet18, FS-EAE improves accuracy by +2.59\% on OpenSARShip and +1.46\% on FUSARShip without APSA. This indicates that the dual-branch structural stabilization effectively enhances robustness against scale variations and domain shifts.

The introduction of APSA further amplifies these advantages. Under the ResNet18+FS-EAE configuration, APSA yields an additional 1.66\% gain on OpenSARShip. Notably, the improvement on the lightweight A-ConvNet significantly outperforms that on the ResNet series, proving that our framework effectively compensates for the limited representational capacity of shallow networks in SAR ATR tasks.

\subsection{Hierarchical Sensitivity Analysis}\label{AA}

We investigate the optimal insertion stage for FS-EAE through the ResNet18 architecture. As reported in Tab.~\ref{tab:ablation_stages}, Layer1 achieved peak performance across all datasets.

\begin{table}[H]
\centering
\caption{results on different stages of FS-EAE. The best results are highlighted in \textbf{bold}.}
\label{tab:ablation_stages}
\normalsize
\resizebox{0.48\textwidth}{!}{%
\begin{tabular}{c|cccc}
\toprule
\textbf{Dataset} & \textbf{Pre (\%)} & \textbf{Layer1 (\%)} & \textbf{Layer2 (\%)} & \textbf{Layer4 (\%)} \\ 
\midrule
OpenSARShip & 74.29 & \textbf{76.42} & 74.76 & 75.94 \\
FUSARShip & 86.41 & \textbf{89.32} & 83.98 & 82.52 \\ 
\bottomrule
\end{tabular}%
}
\end{table}

This phenomenon is intrinsically linked to the physical characteristics of SAR imagery. Early-stage features contain dense frequency-domain information, allowing the wavelet decomposition in FS-EAE to accurately suppress coherent speckle noise. As the network deepens (Layer2 to Layer4), features become highly abstracted. The local multi-scale filters of FS-EAE struggle to parse complex high-level semantic manifolds, and the gradient flow from APSA may suffer from diffusion, leading to the observed performance decline (-6.80\% at Layer4 on FUSARShip).

\subsection{Ablation and Efficiency Analysis}\label{AA}

To comprehensively evaluate the proposed framework, we analyze the contributions of its key components, compare APSA with alternative teacher--student strategies, and assess the computational and runtime efficiency of different branch configurations. These analyses jointly examine the effectiveness, synergy, and practical efficiency of the proposed design.

\begin{table}[htbp]
\centering
\caption{Ablation results on FUSARShip and OpenSARShip datasets.}
\label{tab:ablation_tight}
\footnotesize
\setlength{\tabcolsep}{8pt} 
\begin{tabular}{ccccc}
\toprule
\multicolumn{3}{c}{\textbf{Components}} & \multicolumn{2}{c}{\textbf{Accuracy (\%)}} \\
\cmidrule(lr){1-3} \cmidrule(lr){4-5}
\textbf{Freq.} & \textbf{Spat.} & \textbf{APSA} & \textbf{FUSARShip} & \textbf{OpenSARShip} \\
\midrule
$\times$     & $\times$     & $\times$     & 83.01 & 72.17 \\
$\checkmark$ & $\times$     & $\times$     & 83.98 & 74.53 \\
$\times$     & $\checkmark$ & $\times$     & 83.50 & 74.29 \\
$\times$     & $\times$     & $\checkmark$ & 83.98 & 75.47 \\
$\checkmark$ & $\checkmark$ & $\times$     & 84.47 & 74.76 \\
$\checkmark$ & $\times$     & $\checkmark$ & 85.92 & 73.58 \\
$\times$     & $\checkmark$ & $\checkmark$ & 86.41 & 74.29 \\
$\checkmark$ & $\checkmark$ & $\checkmark$ & \textbf{89.32} & \textbf{76.42} \\
\bottomrule
\end{tabular}
\end{table}

As shown in Table~\ref{tab:ablation_tight}, the ablation results elucidate the complementary roles of feature-mode decoupling and supervision-signal coupling. Introducing the frequency-domain branch alone yields a 0.97\%--2.36\% improvement, confirming that wavelet decomposition effectively captures domain-invariant scattering characteristics. Combining the spatial and frequency branches further improves performance, demonstrating their complementary representation capabilities. In particular, on OpenSARShip, where background clutter is severe, the spatial branch compensates for potential detail loss caused by frequency filtering through long-range contextual modeling. With APSA further incorporated, the full configuration achieves the best performance, reaching 89.32\% and 76.42\% accuracy on FUSARShip and OpenSARShip, respectively. These results indicate that APSA effectively balances spatial and frequency representations through adaptive semantic supervision, guiding the model toward more noise-invariant representations and reducing overfitting to a single feature mode.




\begin{table}[htbp]
\centering
\caption{Comparison of different knowledge distillation strategies on FUSARShip and OpenSARShip. Offline fine-tuned employs a teacher fine-tuned on the target domain; Offline pretrained-frozen freezes the ImageNet-pretrained backbone while learning a randomly initialized task-specific classifier; Offline scratch trains the teacher from random initialization; and Mean Teacher uses an exponential moving average of the student parameters as the teacher.}
\label{tab:distillation_strategy}
\footnotesize
\setlength{\tabcolsep}{5pt}
\renewcommand{\arraystretch}{1.10}
\begin{tabular}{@{}ccc@{}}
\toprule
\textbf{Strategy}
& \textbf{FUSARShip (\%)}
& \textbf{OpenSARShip (\%)} \\
\midrule
Offline fine-tuned
& 86.89 & 72.41 \\
Offline pretrained-frozen
& 84.47 & 74.29 \\
Offline scratch
& 86.41 & 75.47 \\
Mean teacher
& 83.50 & 73.11 \\
\textbf{APSA (Ours)}
& \textbf{89.32} & \textbf{76.42} \\
\bottomrule
\end{tabular}
\end{table}


To further assess APSA, Table~\ref{tab:distillation_strategy} compares it with representative offline and online distillation strategies. APSA achieves the best results, exceeding the strongest alternative by $2.43$ and $0.95$ percentage points on FUSARShip and OpenSARShip, respectively. Mean Teacher performs worse than the offline baselines, suggesting that full-model EMA updates may inherit unreliable student representations during mid-training. APSA instead freezes the pretrained backbone and updates only the classifier, reducing semantic drift while retaining adaptability to the evolving target distribution.

\begin{table}[htbp]
\centering
\caption{Computational complexity and runtime efficiency of different branch configurations. Increments over the corresponding backbone are shown in parentheses. Latency and power are measured on an NVIDIA GeForce RTX 4060 Laptop GPU and reported as the mean $\pm$ standard deviation over five runs.}
\label{tab:complexity_runtime}
\scriptsize
\setlength{\tabcolsep}{2.2pt}
\renewcommand{\arraystretch}{1.10}
\begin{tabular*}{\columnwidth}
{@{\extracolsep{\fill}}ccccc@{}}
\toprule
\makecell[c]{\textbf{Branch}}
& \makecell[c]{\textbf{FLOPs (M)}\\\textbf{($\Delta$FLOPs)}}
& \makecell[c]{\textbf{Params (M)}\\\textbf{($\Delta$Params)}}
& \makecell[c]{\textbf{Latency}\\\textbf{(ms)}}
& \makecell[c]{\textbf{Power}\\\textbf{(W)}} \\
\midrule

\multicolumn{5}{l}{\textit{ResNet18}} \\
Base
& 1823.52 (0.00)
& 11.18 (0.00)
& $7.27{\pm}0.39$
& $17.41{\pm}0.37$ \\
Spatial
& 2367.44 (543.92)
& 11.36 (0.18)
& $11.08{\pm}0.21$
& $15.99{\pm}0.39$ \\
Frequency
& 1830.13 (6.61)
& 11.19 (0.01)
& $11.31{\pm}0.41$
& $14.05{\pm}0.34$ \\
Both
& 2374.05 (550.53)
& 11.37 (0.19)
& $14.36{\pm}0.57$
& $16.75{\pm}0.80$ \\
\midrule

\multicolumn{5}{l}{\textit{ShuffleNetV2-$0.25\times$}} \\
Base
& 22.32 (0.00)
& 0.14 (0.00)
& $16.51{\pm}0.28$
& $10.02{\pm}0.64$ \\
Spatial
& 333.18 (310.86)
& 0.17 (0.03)
& $19.94{\pm}1.54$
& $12.47{\pm}0.08$ \\
Frequency
& 33.00 (10.68)
& 0.15 (0.01)
& $17.70{\pm}3.58$
& $12.79{\pm}1.55$ \\
Both
& 343.87 (321.55)
& 0.17 (0.03)
& $24.11{\pm}0.85$
& $12.56{\pm}0.43$ \\
\bottomrule
\end{tabular*}
\end{table}

Table~\ref{tab:complexity_runtime} summarizes the computational overhead of the proposed branches. The spatial branch accounts for most of the additional FLOPs, while the frequency branch adds only $6.61$M and $10.68$M FLOPs to ResNet18 and ShuffleNetV2-$0.25\times$, respectively. With both branches enabled, FSCE-Net$_{\phi}$ and FSCE-Net$_{\mu}$ require $14.36$ ms and $24.11$ ms per frame, respectively. Notably, FSCE-Net$_{\mu}$ contains only $0.17$M parameters and achieves approximately $41.5$ FPS, demonstrating its suitability for real-time edge deployment.

\subsection{Impact of Convolutional Kernel Sizes}\label{AA}

Furthermore, we investigated various combinations of convolutional kernel sizes within the spatial branch of FS-EAE to validate our structural design. As shown in Tab.~\ref{tab:kernel_size}, the adopted combination ($\{3, 5, 7, 9\}$) achieves the highest accuracy on both datasets. 

\begin{table}[H]
\centering
\caption{Performance comparison of different convolution kernel size combinations.}
\label{tab:kernel_size}
\normalsize
\resizebox{0.48\textwidth}{!}{
\begin{tabular}{ccc} 
\toprule
\multirow{2}{*}{\textbf{Size of convolution kernels}} & \multicolumn{2}{c}{\textbf{Accuracy (\%) $\uparrow$}} \\ \cmidrule(lr){2-3}
& \textbf{OpenSARShip} & \textbf{FUSARShip} \\ 
\midrule
\boldmath{$3\times3, 5\times5, 7\times7, 9\times9$} & \textbf{76.42} & \textbf{89.32} \\
$3\times3, 5\times5, 7\times7, 11\times11$ & 75.24 & 84.95 \\
$3\times3, 5\times5, 9\times9, 11\times11$ & 73.82 & 84.47 \\
$3\times3, 7\times7, 9\times9, 11\times11$ & 73.35 & 83.98 \\
$5\times5, 7\times7, 9\times9, 11\times11$ & 75.47 & 85.92 \\
$3\times3, 5\times5, 7\times7$ & 74.76 & 84.95 \\
$5\times5, 7\times7, 9\times9$ & 73.82 & 84.47 \\
$3\times3, 7\times7, 11\times11$ & 75.24 & 86.41 \\ 
\bottomrule
\end{tabular}%
}
\end{table}

These results confirm that the selected kernels enable effective multi-scale representation, capturing everything from fine-grained scattering textures to global target structures. In contrast, introducing an $11\times11$ kernel leads to performance degradation, likely due to the over-inclusion of background clutter and a mismatch with the intrinsic scale of ship targets. Similarly, reducing kernel diversity limits the model’s capacity to capture scale-dependent features, resulting in inferior accuracy. These findings justify the rationality of the multi-scale configuration in our FS-EAE module.

\subsection{APSA Hyperparameter Analysis}\label{AA}
\begin{figure}[h]
\centering
\includegraphics[width=0.45\textwidth]{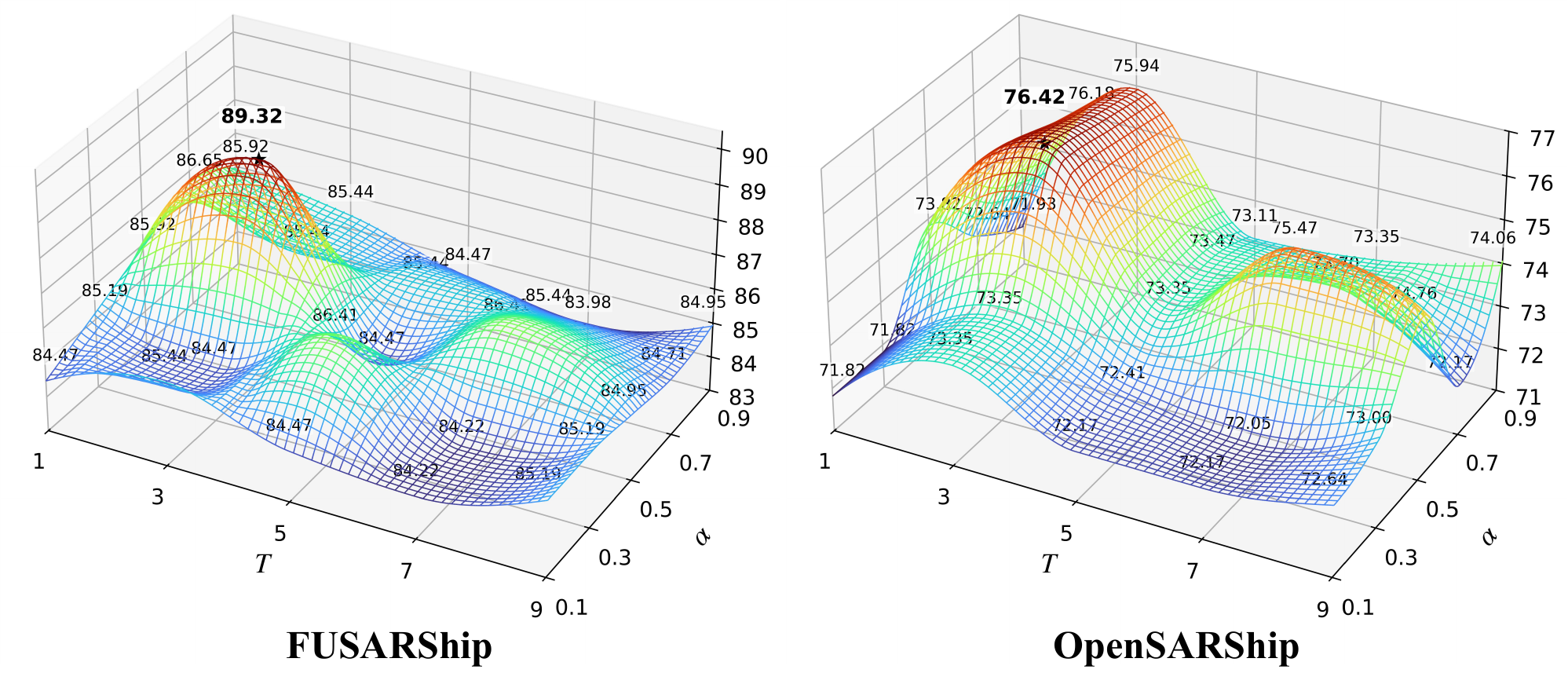}
\caption{Joint sensitivity of $\alpha$ and $T$. Labels denote measured validation accuracies; surfaces show separable 2-D PCHIP interpolation~\cite{fritsch1984method}.}
\label{fig:hyperparameter_surface}
\end{figure}

We further investigate the joint influence of the temperature $T$ and loss weight $\alpha$ on APSA through a grid search over $T\in\{1,3,5,7,9\}$ and $\alpha\in\{0.1,0.3,0.5,0.7,0.9\}$. The quantitative results are reported in Table~\ref{tab:hyperparameter_grid}, while Fig.~\ref{fig:hyperparameter_surface} visualizes the corresponding response surfaces. Despite the different data characteristics of FUSARShip and OpenSARShip, both datasets consistently achieve their best validation accuracy at $T=3$ and $\alpha=0.5$, reaching 89.32\% and 76.42\%, respectively. Performance generally decreases when either hyperparameter deviates from this configuration, indicating that APSA benefits from a moderate temperature and a balanced supervision weight.

\begin{table}[h]
\centering
\caption{Joint grid search of $\alpha$ and $T$. Validation accuracy (\%) is reported for each hyperparameter combination, with the best result on each dataset highlighted in bold.}
\label{tab:hyperparameter_grid}
\small
\setlength{\tabcolsep}{5.2pt}
\renewcommand{\arraystretch}{1.10}

\begin{tabular}{c|ccccc}
\toprule
\multicolumn{6}{c}{\textbf{FUSARShip}} \\
\midrule
$\alpha \backslash T$
& 1 & 3 & 5 & 7 & 9 \\
\midrule
0.1 & 84.47 & 85.44 & 84.47 & 84.22 & 85.19 \\
0.3 & 85.19 & 84.47 & 86.41 & 84.22 & 85.19 \\
0.5 & 85.92 & \textbf{89.32} & 84.47 & 86.41 & 84.95 \\
0.7 & 86.65 & 85.44 & 85.44 & 85.44 & 84.71 \\
0.9 & 85.92 & 85.44 & 84.47 & 83.98 & 84.95 \\
\midrule
\multicolumn{6}{c}{\textbf{OpenSARShip}} \\
\midrule
$\alpha \backslash T$
& 1 & 3 & 5 & 7 & 9 \\
\midrule
0.1 & 71.82 & 73.35 & 72.17 & 72.17 & 72.64 \\
0.3 & 71.82 & 73.35 & 72.41 & 72.05 & 73.00 \\
0.5 & 73.82 & \textbf{76.42} & 73.35 & 75.47 & 74.76 \\
0.7 & 72.64 & 76.18 & 73.47 & 73.70 & 72.17 \\
0.9 & 71.93 & 75.94 & 73.11 & 73.35 & 74.06 \\
\bottomrule
\end{tabular}
\end{table}

The observed optimum can be attributed to the complementary roles of the two hyperparameters. A moderate temperature ($T=3$) softens the teacher prediction distribution sufficiently to expose inter-class semantic relationships while avoiding excessive smoothing at larger temperatures. Meanwhile, $\alpha=0.5$ provides an appropriate balance between teacher-guided semantic supervision and ground-truth discrimination, allowing the student to acquire robust semantic knowledge without being dominated by either supervisory signal. These consistent trends across both datasets support the use of $T=3$ and $\alpha=0.5$ as the default APSA configuration in all subsequent experiments.



\section{Conclusion}

In this paper, we proposed the FSCE framework, which synergistically integrates the FS-EAE module and the APSA mechanism to address speckle-induced feature fragmentation in SAR ATR. By combining early-stage dual-domain filtering with on-policy semantic alignment, the framework improves representation robustness under noisy SAR conditions. Extensive experiments demonstrate that $\text{FSCE-Net}_{\Phi}$ achieves superior performance, while the lightweight $\text{FSCE-Net}_{\mu}$ (0.17M parameters) provides superior efficiency for edge deployment. Although performance under extreme scale variations warrants further improvement, the proposed framework establishes a robust and theoretically grounded paradigm for complex SAR environments. Future work will investigate scale-aware representations for cross-resolution robustness and extend the underlying noise--structure decoupling principle to related noise-robust target perception tasks.


\section*{Acknowledgments}
This work was supported by the Sichuan Science and Technology Program (granted No. 2024ZDZX0011, No. 2026NSFSC1482 and No. 2025ZHCG0002).

\newpage


\begin{IEEEbiography}[{\includegraphics[width=1in,height=1.25in,clip,keepaspectratio]{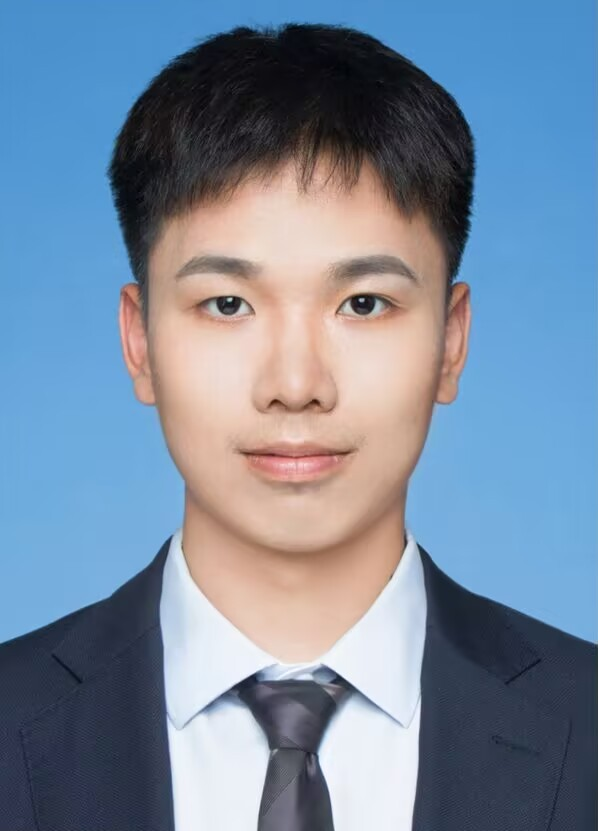}}]{Yansong Lin}
 received the B.E. degree from Wuhan Institute of Technology in 2025. He is currently pursuing the M.S. degree at the University of Electronic Science and Technology of China (UESTC). His research interests include knowledge distillation and AI security.\end{IEEEbiography}

\begin{IEEEbiography}[{\includegraphics[width=1in,height=1.25in,clip,keepaspectratio]{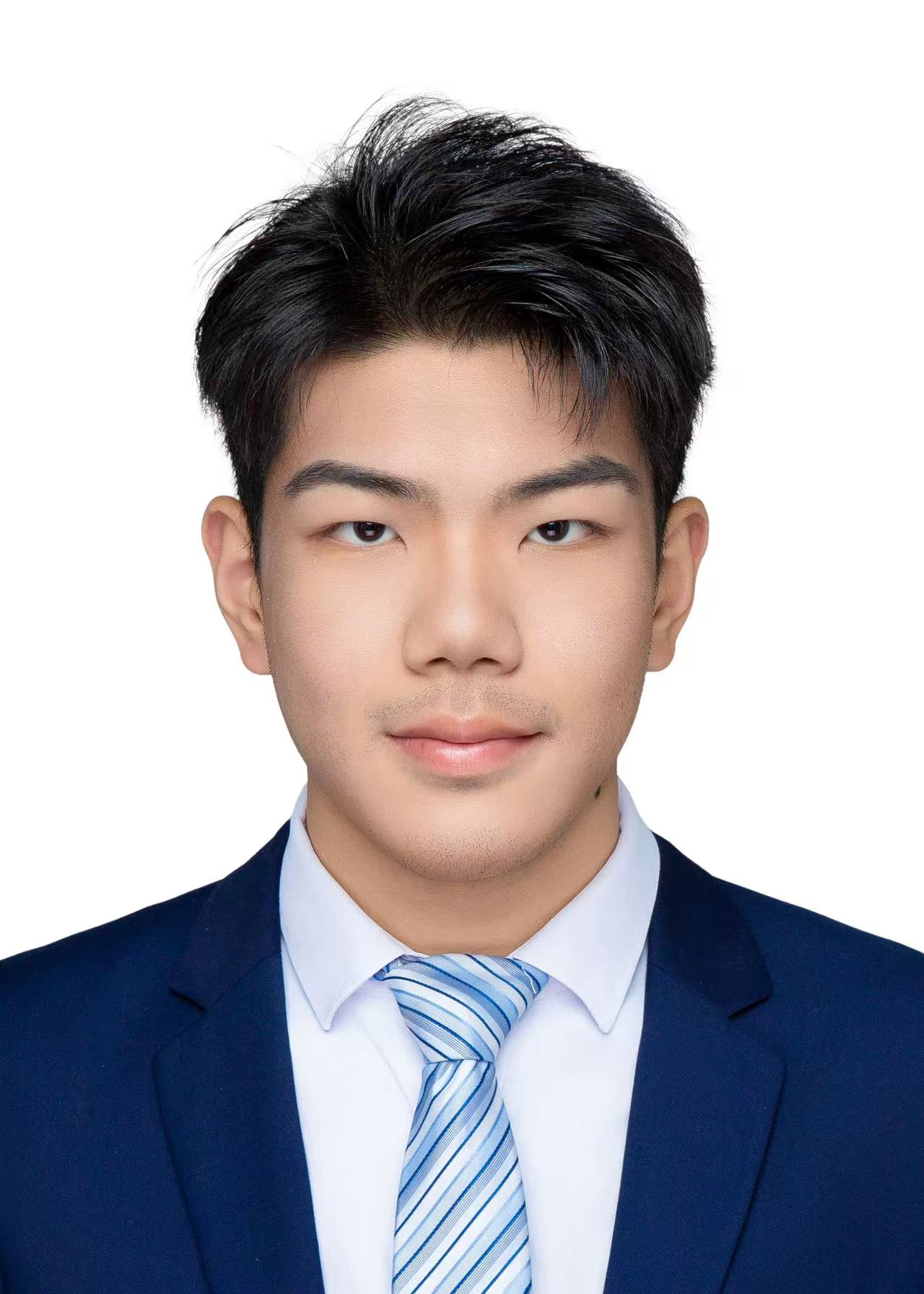}}]{Zihan Cheng}
 received his B.E. from the University of Electronic Science and Technology of China (UESTC) in 2023. He is currently majoring in computer science as a master's candidate at UESTC. His research interests include deep learning and knowledge distillation.\end{IEEEbiography}

\begin{IEEEbiography}[{\includegraphics[width=1in,height=1.25in,clip,keepaspectratio]{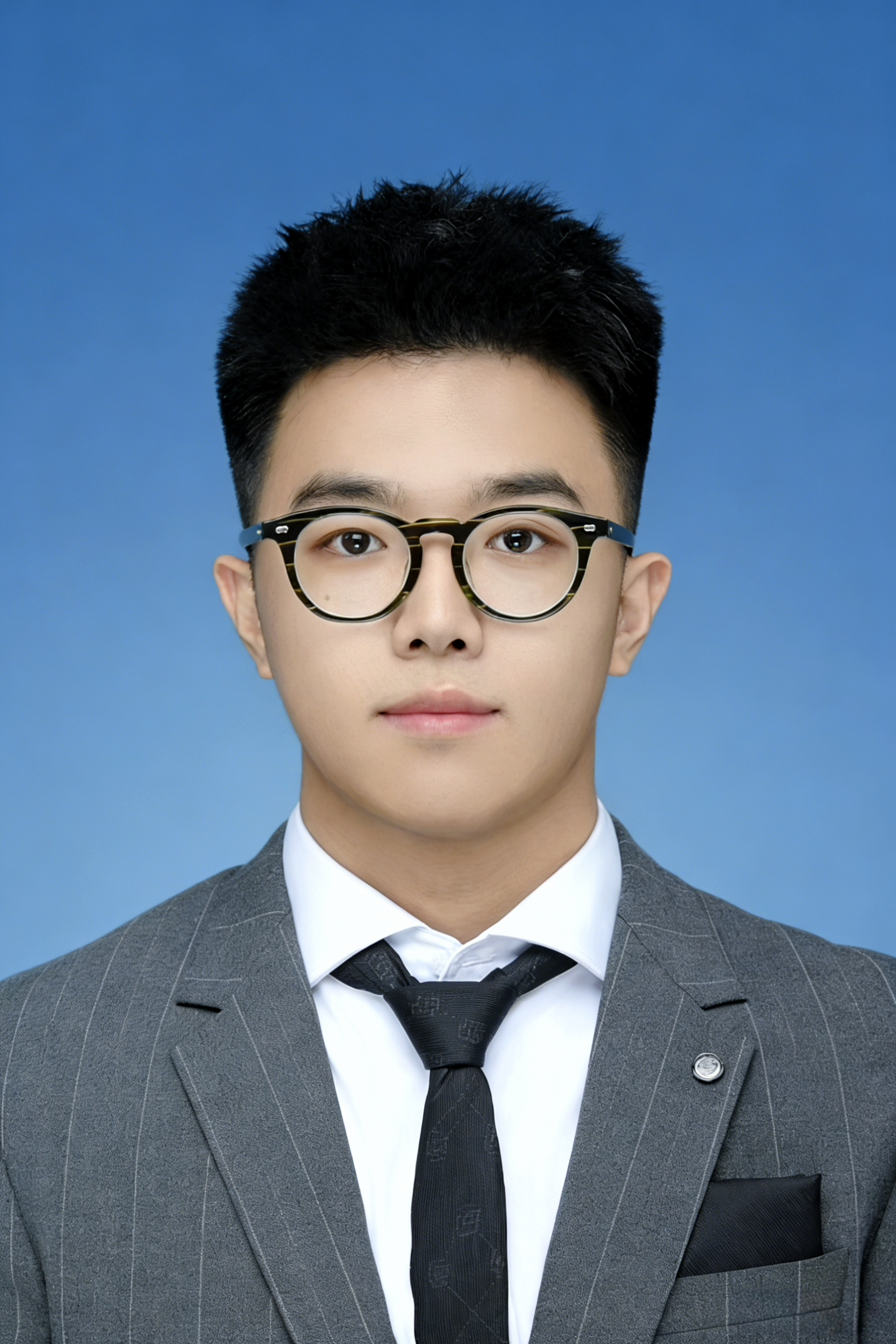}}]{Ziyue Yang}
 is currently a junior student in Intelligent Science and Technology at Wuhan Institute of Technology. His research interests include computer vision and image fusion.\end{IEEEbiography}

 \begin{IEEEbiography}[{\includegraphics[width=1in,height=1.25in,clip,keepaspectratio]{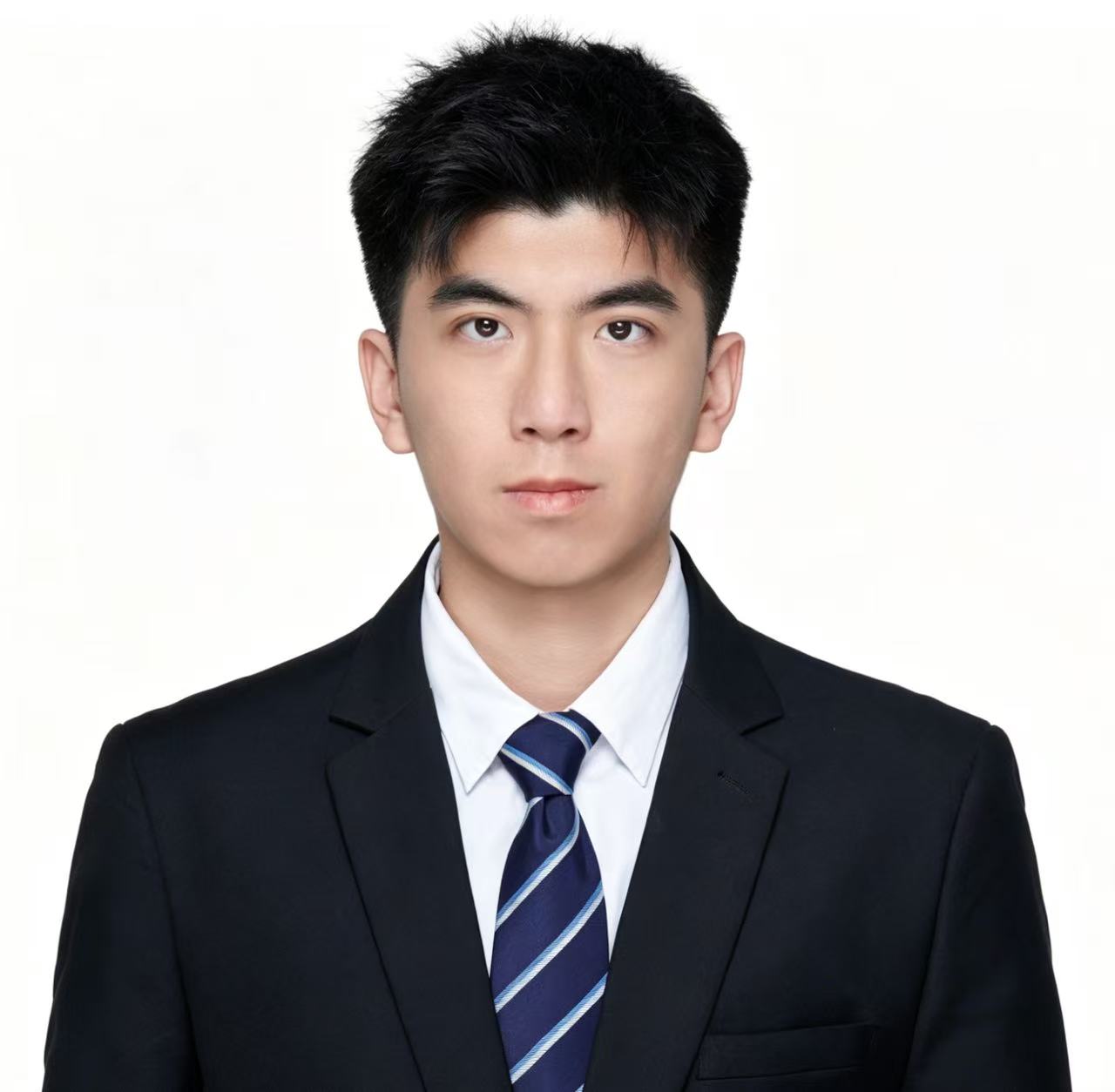}}]{Xinming Wang}
  received the B.E. degree from Tianjin University in 2024. He is currently working toward the PhD degree in pattern recognition and intelligent systems from the Institute of Automation, Chinese Academy of Sciences. His research interests include LLM agents and reinforcement learning.\end{IEEEbiography}

\begin{IEEEbiography}[{\includegraphics[width=1in,height=1.25in,clip,keepaspectratio]{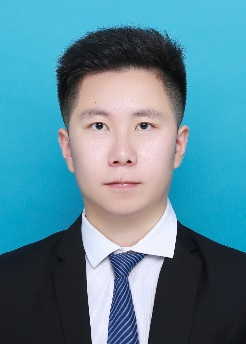}}]{Jielei Wang}
 received his B.E. degree from Chongqing University of Posts and Telecommunications (CQUPT), Chongqing, China, in 2018, and his Ph.D. degree from the University of Electronic Science and Technology of China (UESTC), Chengdu, China, in 2023.
 
 He is currently a Postdoctoral Research Fellow with the Institute of Intelligent Computing, UESTC. His research interests include synthetic aperture radar (SAR) image processing, AI security, image super-resolution, deep neural network compression, and edge computing.\end{IEEEbiography}

\begin{IEEEbiography}[{\includegraphics[width=1in]{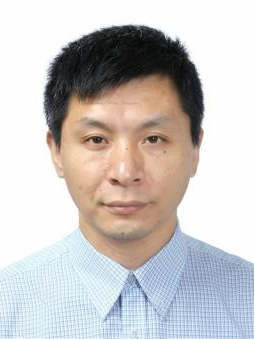}}]{Guoming Lu}
 received his Ph.D. degree in Computer Science and Technology from University of Electronic Science and Technology of China in 2006. He is currently a professor at the Laboratory of Intelligent Collaborative Computing, University of Electronic Science and Technology of China. His research interests include Heterogeneous Computing.\end{IEEEbiography}

\begin{IEEEbiography}[{\includegraphics[width=1in,height=1.25in,clip,keepaspectratio]{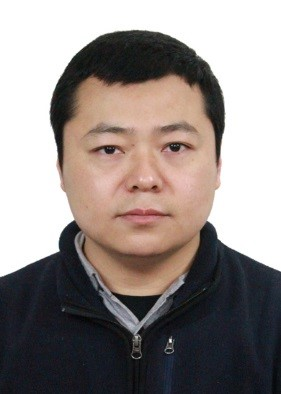}}]{Zongyong Cui}
 (S’13–M’15) received the B.E. and Ph.D. degrees in signal and information processing from the University of Electronic Science and Technology of China (UESTC), Chengdu, China, in 2007 and 2015, respectively. From 2013 to 2014, he was a Visiting Student with the Department of Electrical and Computer Engineering, National University of Singapore, Singapore.
 
 He is currently a Professor with the School of Information and Communication Engineering in UESTC. His research interests include radar image processing, target recognition and machine learning. 
 
 Dr. Cui is a Reviewer of the IEEE GEOSCIENCE AND REMOTE SENSING LETTERS, the IET Radar, Sonar \& Navigation, Journal of Electronic Imaging, and so on.\end{IEEEbiography}


\begin{thebibliography}{10}
\providecommand{\url}[1]{#1}
\csname url@samestyle\endcsname
\providecommand{\newblock}{\relax}
\providecommand{\bibinfo}[2]{#2}
\providecommand{\BIBentrySTDinterwordspacing}{\spaceskip=0pt\relax}
\providecommand{\BIBentryALTinterwordstretchfactor}{4}
\providecommand{\BIBentryALTinterwordspacing}{\spaceskip=\fontdimen2\font plus
\BIBentryALTinterwordstretchfactor\fontdimen3\font minus
  \fontdimen4\font\relax}
\providecommand{\BIBforeignlanguage}[2]{{%
\expandafter\ifx\csname l@#1\endcsname\relax
\typeout{** WARNING: IEEEtran.bst: No hyphenation pattern has been}%
\typeout{** loaded for the language `#1'. Using the pattern for}%
\typeout{** the default language instead.}%
\else
\language=\csname l@#1\endcsname
\fi
#2}}
\providecommand{\BIBdecl}{\relax}
\BIBdecl

\bibitem{cigna2024exploiting}
F.~Cigna, T.~Balz, D.~Tapete, G.~Caspari, B.~Fu, M.~Abballe, and H.~Jiang,
  ``Exploiting satellite sar for archaeological prospection and heritage site
  protection,'' \emph{Geo-spatial Information Science}, vol.~27, no.~3, pp.
  526--551, 2024.

\bibitem{sunantha2025machine}
O.~Sunantha, Z.~Shao, P.~Pattama, A.~Potchara, X.~Huang, and A.~Zeeshan,
  ``Machine learning-based estimation of soil organic carbon in thailand’s
  cash crops using multispectral and sar data fusion combined with
  environmental variables,'' \emph{Geo-spatial Information Science}, pp. 1--23,
  2025.

\bibitem{li2025video}
M.~Li, Z.~Wu, W.~Huo, J.~Wu, and J.~Zhang, ``Video sar image reconstruction
  based on sparse tensor recovery and unfolded transformer,'' \emph{IEEE
  Transactions on Circuits and Systems for Video Technology}, 2025.

\bibitem{wei2024displacements}
L.~Wei, F.~Wang, C.~Tolomei, S.~Liu, C.~Bignami, B.~Li, D.~Lv, E.~Trasatti,
  Y.~Cui, G.~Ventura \emph{et~al.}, ``Displacements of fushun west opencast
  coal mine revealed by multi-temporal insar technology,'' \emph{Geo-Spatial
  Information Science}, vol.~27, no.~3, pp. 585--601, 2024.

\bibitem{chen2025mdca}
T.~Chen, J.~Pan, J.~Xu, J.~Hu, L.~Cao, and J.~Li, ``Mdca-net: a
  multi-directional alignment and dynamic context aggregation network for
  optical and sar image fusion,'' \emph{Geo-spatial Information Science}, pp.
  1--19, 2025.

\bibitem{elyouncha2024synergistic}
A.~Elyouncha, G.~Brostr{\"o}m, and H.~Johnsen, ``Synergistic utilization of
  spaceborne sar observations for monitoring the baltic sea flow through the
  danish straits,'' \emph{Earth and Space Science}, vol.~11, no.~10, p.
  e2024EA003794, 2024.

\bibitem{huang2024waterdetectionnet}
B.~Huang, P.~Li, H.~Lu, J.~Yin, Z.~Li, and H.~Wang, ``Waterdetectionnet: a new
  deep learning method for flood mapping with sar image convolutional neural
  network,'' \emph{IEEE Journal of Selected Topics in Applied Earth
  Observations and Remote Sensing}, 2024.

\bibitem{jiao2025glacier}
H.~Jiao, G.~Li, Z.~Chen, and X.~Cheng, ``Glacier surface melt monitoring using
  sentinel-1 sar backscattering coefficient and polarimetric decomposition
  features at greenland ice sheet,'' \emph{Geo-spatial Information Science},
  pp. 1--23, 2025.

\bibitem{bai2025investigation}
Z.~Bai, Y.~Qin, Y.~Wang, B.~Yang, Y.~Lin, M.~Li, G.~Hao, Y.~Li, and W.~Shen,
  ``An investigation into multi-scale periodic dynamic deformation monitoring
  of bridges through integrated spaceborne and ground-based insar,''
  \emph{Geo-spatial Information Science}, pp. 1--16, 2025.

\bibitem{maharjan2026amplitude}
P.~Maharjan, Z.~Li, and C.~McGuiness, ``Amplitude, phase, and gradient recovery
  from compressed sar images,'' \emph{IEEE Transactions on Circuits and Systems
  for Video Technology}, 2026.

\bibitem{huang2025physics}
Z.~Huang, L.~Liu, S.~Yang, Z.~Wang, G.~Cheng, and J.~Han, ``Physics-guided
  detector for sar airplanes,'' \emph{IEEE Transactions on Circuits and Systems
  for Video Technology}, 2025.

\bibitem{wang2025rethinking}
J.~Wang, X.~Chen, X.~Huang, R.~Zhang, Y.~Wang, and T.~Lu, ``Rethinking the role
  of panchromatic images in pan-sharpening,'' \emph{IEEE Transactions on
  Multimedia}, 2025.

\bibitem{dong2025complex}
G.~Dong, Y.~Wang, H.~Liu, and S.~Liu, ``Complex-valued sar image
  super-resolution via sub-aperture learning and fusion,'' \emph{IEEE
  Transactions on Geoscience and Remote Sensing}, 2025.

\bibitem{li2021multiscale}
Y.~Li, L.~Du, and D.~Wei, ``Multiscale cnn based on component analysis for sar
  atr,'' \emph{IEEE Transactions on Geoscience and Remote Sensing}, vol.~60,
  pp. 1--12, 2021.

\bibitem{feng2022electromagnetic}
S.~Feng, K.~Ji, F.~Wang, L.~Zhang, X.~Ma, and G.~Kuang, ``Electromagnetic
  scattering feature (esf) module embedded network based on asc model for
  robust and interpretable sar atr,'' \emph{IEEE Transactions on Geoscience and
  Remote Sensing}, vol.~60, pp. 1--15, 2022.

\bibitem{huang2021physics}
Z.~Huang, C.~O. Dumitru, and J.~Ren, ``Physics-aware feature learning of sar
  images with deep neural networks: A case study,'' in \emph{2021 IEEE
  International Geoscience and Remote Sensing Symposium IGARSS}.\hskip 1em plus
  0.5em minus 0.4em\relax IEEE, 2021, pp. 1264--1267.

\bibitem{qin2024scattering}
J.~Qin, B.~Zou, Y.~Chen, H.~Li, and L.~Zhang, ``Scattering attribute embedded
  network for few-shot sar atr,'' \emph{IEEE Transactions on Aerospace and
  Electronic Systems}, 2024.

\bibitem{miao2024time}
Z.~Miao and M.~Zhao, ``Time--space--frequency feature fusion for 3-channel
  motor imagery classification,'' \emph{Biomedical Signal Processing and
  Control}, vol.~90, p. 105867, 2024.

\bibitem{wang2025forgotten}
J.~Wang, Y.~Lin, C.~Chen, X.~Huang, R.~Zhang, Y.~Wang, and T.~Lu, ``From
  forgotten to pan-sharpening,'' \emph{Pattern Recognition}, p. 112653, 2025.

\bibitem{cheng2014sar}
J.~Cheng, L.~Li, H.~Li, and F.~Wang, ``Sar target recognition based on improved
  joint sparse representation,'' \emph{EURASIP Journal on Advances in Signal
  Processing}, vol. 2014, pp. 1--12, 2014.

\bibitem{lin2013optimizing}
C.~Lin, B.~Wang, X.~Zhao, and M.~Pang, ``Optimizing kernel pca using sparse
  representation-based classifier for mstar sar image target recognition,''
  \emph{Mathematical Problems in Engineering}, vol. 2013, no.~1, p. 847062,
  2013.

\bibitem{chen2016target}
S.~Chen, H.~Wang, F.~Xu, and Y.-Q. Jin, ``Target classification using the deep
  convolutional networks for sar images,'' \emph{IEEE transactions on
  geoscience and remote sensing}, vol.~54, no.~8, pp. 4806--4817, 2016.

\bibitem{zhao2022few}
X.~Zhao, X.~Lv, J.~Cai, J.~Guo, Y.~Zhang, X.~Qiu, and Y.~Wu, ``Few-shot sar-atr
  based on instance-aware transformer,'' \emph{Remote Sensing}, vol.~14, no.~8,
  p. 1884, 2022.

\bibitem{kim2024synthetic}
M.~Kim and S.~Kim, ``Synthetic sar data domain randomization for unseen sar
  atr,'' in \emph{Algorithms for Synthetic Aperture Radar Imagery XXXI}, vol.
  13032.\hskip 1em plus 0.5em minus 0.4em\relax SPIE, 2024, pp. 180--184.

\bibitem{gao2024asc}
Y.~Gao, W.~Guo, D.~Li, and W.~Yu, ``Asc-rise: Physical information guided
  explanation of sar atr models,'' in \emph{IGARSS 2024-2024 IEEE International
  Geoscience and Remote Sensing Symposium}.\hskip 1em plus 0.5em minus
  0.4em\relax IEEE, 2024, pp. 2159--2162.

\bibitem{ranchin1993wavelet}
T.~Ranchin and L.~Wald, ``The wavelet transform for the analysis of remotely
  sensed images,'' \emph{International Journal of Remote Sensing}, vol.~14,
  no.~3, pp. 615--619, 1993.

\bibitem{song2023fourier}
B.~Song, S.~Min, H.~Yang, Y.~Wu, and B.~Wang, ``A fourier frequency domain
  convolutional neural network for remote sensing crop classification
  considering global consistency and edge specificity,'' \emph{Remote Sensing},
  vol.~15, no.~19, p. 4788, 2023.

\bibitem{zi2023wavelet}
Y.~Zi, H.~Ding, F.~Xie, Z.~Jiang, and X.~Song, ``Wavelet integrated
  convolutional neural network for thin cloud removal in remote sensing
  images,'' \emph{Remote Sensing}, vol.~15, no.~3, p. 781, 2023.

\bibitem{he2019lifting}
C.~He, Z.~Shi, T.~Qu, D.~Wang, and M.~Liao, ``Lifting scheme-based deep neural
  network for remote sensing scene classification,'' \emph{Remote Sensing},
  vol.~11, no.~22, p. 2648, 2019.

\bibitem{dang2024dctransformer}
Y.~Dang, X.~Zhang, H.~Zhao, and B.~Liu, ``Dctransformer: A channel attention
  combined discrete cosine transform to extract spatial--spectral feature for
  hyperspectral image classification,'' \emph{Applied Sciences}, vol.~14,
  no.~5, p. 1701, 2024.

\bibitem{li20253d}
G.~Li and M.~Ye, ``3d wavelet convolutions with extended receptive fields for
  hyperspectral image classification,'' \emph{arXiv preprint arXiv:2504.10795},
  2025.

\bibitem{wang2025hitchhiker}
X.~Wang, J.~Xu, A.~H. Feng, Y.~Chen, H.~Guo, F.~Zhu, Y.~Shao, M.~Ren, H.~Yi,
  S.~Lian \emph{et~al.}, ``The hitchhiker’s guide to autonomous research: A
  survey of scientific agents,'' \emph{TechRxiv.August 07, 2025.
  DOI:10.36227/techrxiv175459840.02185500/V1}, 2025.

\bibitem{liu2026atrnet}
Y.~Liu, W.~Li, L.~Liu, J.~Zhou, B.~Peng, Y.~Song, X.~Xiong, W.~Yang, T.~Liu,
  Z.~Liu \emph{et~al.}, ``Atrnet-star: A large dataset and benchmark towards
  remote sensing object recognition in the wild,'' \emph{IEEE Transactions on
  Pattern Analysis and Machine Intelligence}, 2026.

\bibitem{wang2026beyond}
X.~Wang, W.~Wang, H.~Yang, Y.~Lin, Z.~Ruan, S.~Peng, Q.~Peng, N.~Qiao, F.~Lu,
  G.~Ma \emph{et~al.}, ``Beyond correctness: Benchmarking and aligning response
  behaviors in hybrid-thinking mllms,'' \emph{arXiv preprint arXiv:2608.12781},
  2026.

\bibitem{wang2025mr}
X.~Wang, J.~Xu, B.~Yu, S.~Lian, H.~Yi, Y.~Chen, Y.~Zhu, B.~Wang, H.~Yang, H.~Hu
  \emph{et~al.}, ``Mr-align: Meta-reasoning informed factuality alignment for
  large reasoning models,'' \emph{arXiv preprint arXiv:2510.24794}, 2025.

\bibitem{guo2026baddenoise}
J.~Guo, Y.~Lin, M.~Jiang, J.~Wang, C.~Chen, W.~Jiang, and H.~Li, ``Baddenoise:
  Backdoor attacks on self-supervised image denoising,'' \emph{Pattern
  Recognition}, p. 114565, 2026.

\bibitem{yang2023online}
C.~Yang, Z.~An, H.~Zhou, F.~Zhuang, Y.~Xu, and Q.~Zhang, ``Online knowledge
  distillation via mutual contrastive learning for visual recognition,''
  \emph{IEEE Transactions on Pattern Analysis and Machine Intelligence},
  vol.~45, no.~8, pp. 10\,212--10\,227, 2023.

\bibitem{guo2020online}
Q.~Guo, X.~Wang, Y.~Wu, Z.~Yu, D.~Liang, X.~Hu, and P.~Luo, ``Online knowledge
  distillation via collaborative learning,'' in \emph{Proceedings of the
  IEEE/CVF Conference on Computer Vision and Pattern Recognition}, 2020, pp.
  11\,020--11\,029.

\bibitem{song2023efficient}
H.~Song, C.~Wei, and Z.~Yong, ``Efficient knowledge distillation for remote
  sensing image classification: a cnn-based approach,'' \emph{International
  Journal of Web Information Systems}, vol.~20, no.~2, pp. 129--158, 2023.

\bibitem{le2023knowledge}
H.-{\^A}. L{\^e} and M.-T. Pham, ``Knowledge distillation for object detection:
  from generic to remote sensing datasets,'' in \emph{IGARSS 2023-2023 IEEE
  International Geoscience and Remote Sensing Symposium}.\hskip 1em plus 0.5em
  minus 0.4em\relax IEEE, 2023, pp. 6194--6197.

\bibitem{keydel1996mstar}
E.~R. Keydel, S.~W. Lee, and J.~T. Moore, ``Mstar extended operating
  conditions: A tutorial,'' \emph{Algorithms for synthetic aperture radar
  imagery III}, vol. 2757, pp. 228--242, 1996.

\bibitem{hou2020fusar}
X.~Hou, W.~Ao, Q.~Song, J.~Lai, H.~Wang, and F.~Xu, ``Fusar-ship: Building a
  high-resolution sar-ais matchup dataset of gaofen-3 for ship detection and
  recognition,'' \emph{Science China Information Sciences}, vol.~63, pp. 1--19,
  2020.

\bibitem{li2017opensarship}
B.~Li, B.~Liu, L.~Huang, W.~Guo, Z.~Zhang, and W.~Yu, ``Opensarship 2.0: A
  large-volume dataset for deeper interpretation of ship targets in sentinel-1
  imagery,'' in \emph{2017 SAR in Big Data Era: Models, Methods and
  Applications (BIGSARDATA)}.\hskip 1em plus 0.5em minus 0.4em\relax IEEE,
  2017, pp. 1--5.

\bibitem{tan2019efficientnet}
M.~Tan and Q.~Le, ``Efficientnet: Rethinking model scaling for convolutional
  neural networks,'' in \emph{International conference on machine
  learning}.\hskip 1em plus 0.5em minus 0.4em\relax PMLR, 2019, pp. 6105--6114.

\bibitem{szegedy2016rethinking}
C.~Szegedy, V.~Vanhoucke, S.~Ioffe, J.~Shlens, and Z.~Wojna, ``Rethinking the
  inception architecture for computer vision,'' in \emph{Proceedings of the
  IEEE conference on computer vision and pattern recognition}, 2016, pp.
  2818--2826.

\bibitem{huang2017densely}
G.~Huang, Z.~Liu, L.~Van Der~Maaten, and K.~Q. Weinberger, ``Densely connected
  convolutional networks,'' in \emph{Proceedings of the IEEE conference on
  computer vision and pattern recognition}, 2017, pp. 4700--4708.

\bibitem{woo2023convnext}
S.~Woo, S.~Debnath, R.~Hu, X.~Chen, Z.~Liu, I.~S. Kweon, and S.~Xie, ``Convnext
  v2: Co-designing and scaling convnets with masked autoencoders,'' in
  \emph{Proceedings of the IEEE/CVF conference on computer vision and pattern
  recognition}, 2023, pp. 16\,133--16\,142.

\bibitem{heo2024rotary}
B.~Heo, S.~Park, D.~Han, and S.~Yun, ``Rotary position embedding for vision
  transformer,'' in \emph{European Conference on Computer Vision}.\hskip 1em
  plus 0.5em minus 0.4em\relax Springer, 2024, pp. 289--305.

\bibitem{chu2021twins}
X.~Chu, Z.~Tian, Y.~Wang, B.~Zhang, H.~Ren, X.~Wei, H.~Xia, and C.~Shen,
  ``Twins: Revisiting the design of spatial attention in vision transformers,''
  \emph{Advances in neural information processing systems}, vol.~34, pp.
  9355--9366, 2021.

\bibitem{ma2024rewrite}
X.~Ma, X.~Dai, Y.~Bai, Y.~Wang, and Y.~Fu, ``Rewrite the stars,'' in
  \emph{Proceedings of the IEEE/CVF conference on computer vision and pattern
  recognition}, 2024, pp. 5694--5703.

\bibitem{yuan2022volo}
L.~Yuan, Q.~Hou, Z.~Jiang, J.~Feng, and S.~Yan, ``Volo: Vision outlooker for
  visual recognition,'' \emph{IEEE transactions on pattern analysis and machine
  intelligence}, vol.~45, no.~5, pp. 6575--6586, 2022.

\bibitem{chi2020fast}
L.~Chi, B.~Jiang, and Y.~Mu, ``Fast fourier convolution,'' \emph{Advances in
  neural information processing systems}, vol.~33, pp. 4479--4488, 2020.

\bibitem{wang2019sar}
L.~Wang, X.~Bai, and F.~Zhou, ``Sar atr of ground vehicles based on esenet,''
  \emph{Remote Sensing}, vol.~11, no.~11, p. 1316, 2019.

\bibitem{zeng2021unknown}
Z.~Zeng, J.~Sun, C.~Xu, and H.~Wang, ``Unknown sar target identification method
  based on feature extraction network and kld--rpa joint discrimination,''
  \emph{Remote Sensing}, vol.~13, no.~15, p. 2901, 2021.

\bibitem{huang2020classification}
Z.~Huang, C.~O. Dumitru, Z.~Pan, B.~Lei, and M.~Datcu, ``Classification of
  large-scale high-resolution sar images with deep transfer learning,''
  \emph{IEEE Geoscience and Remote Sensing Letters}, vol.~18, no.~1, pp.
  107--111, 2020.

\bibitem{choi2022fusion}
J.-H. Choi, M.-J. Lee, N.-H. Jeong, G.~Lee, and K.-T. Kim, ``Fusion of target
  and shadow regions for improved sar atr,'' \emph{IEEE Transactions on
  Geoscience and Remote Sensing}, vol.~60, pp. 1--17, 2022.

\bibitem{jingyi2025target}
C.~Jingyi, Z.~Yang, Y.~Ya’nan, W.~Yamin, Y.~Feng, R.~Weijia, and L.~Jun,
  ``Target recognition method based on graph structure perception of invariant
  features for sar images,'' \emph{Journal of Radars}, vol.~14, no.~2, pp.
  366--388, 2025, doi: 10.12000/JR24125.

\bibitem{wang2024multiresolution}
L.~Wang, Y.~Yang, and Z.~Liu, ``Multiresolution sar target recognition based on
  physical attention enhancement and scale distillation,'' \emph{IEEE
  Transactions on Aerospace and Electronic Systems}, vol.~60, no.~3, pp.
  3081--3094, 2024.

\bibitem{rw2019timm}
R.~Wightman, ``Pytorch image models,''
  \url{https://github.com/rwightman/pytorch-image-models}, 2019.

\bibitem{fritsch1984method}
F.~N. Fritsch and J.~Butland, ``A method for constructing local monotone
  piecewise cubic interpolants,'' \emph{SIAM journal on scientific and
  statistical computing}, vol.~5, no.~2, pp. 300--304, 1984.

\end{thebibliography}
\end{document}